\documentclass[10pt,a4paper]{article}

\usepackage[utf8]{inputenc}
\usepackage[T1]{fontenc}
\usepackage{lmodern}
\usepackage{amsmath,amssymb}
\usepackage{graphicx}
\usepackage{booktabs}
\usepackage[hidelinks]{hyperref}
\usepackage{microtype}
\usepackage[dvipsnames]{xcolor}
\definecolor{refblue}{HTML}{3184BE}
\usepackage[margin=2.2cm]{geometry}
\usepackage{cite}
\usepackage{caption}
\usepackage{subcaption}
\usepackage{float}
\usepackage{enumitem}
\usepackage{tikz}
\usetikzlibrary{positioning,calc}

\setlist{nosep,leftmargin=*}
\title{\textbf{Emergent Intelligence:} \\
Resonant Oscillators Produce Proactive Adaptive Behavior}
\author{Alex Fedosov\thanks{Corresponding author: alex\_fedosov@foundaition.org}, Maxim Yakimenko, Sander Stepanov \\
\textit{FoundAItion Inc.}}
\date{}

\begin{document}

\maketitle

\begin{abstract}
Most artificial neural systems are optimized to transform presented inputs into outputs. Adaptive agents face a prior problem: they must act in the absence of sufficient evidence, generate encounters with the world, and reorganize behavior when evidence appears. We propose a different starting point for intelligent neural networks: proactive search in the absence of signals, curiosity in its most basic form. We investigate whether that capacity can arise from a minimal untrained circuit.  The spiking unit studied here inverts its response to input: with no signal inside its temporal window it fires at higher rates, and once signals are detected it switches to a slower, redirected output regime. Search requires three or more such oscillators, wired in counter-phase, each evaluating the same input over a different temporal window. With no training, supervision, parameter adaptation, or explicit controller, the composite switches autonomously between exploratory spiral search and exploitative tracking, identifying both first-degree spatial symmetry and second-degree spatial groupings. The switch is driven by temporal disagreement between its fast and slow evaluations of the same signal. We view the circuit as evolutionary trained: its capabilities are shaped by structural constraints rather than learned from experience.  Ablation across 63 configurations and 63,000 trials shows the switch requires both temporal staggering and counter-phase opposition, neither of which suffices alone, identifying the behavior as emergent rather than programmed. The spiral persists at zero rotational diffusion, so it is structural, and it degrades gracefully under parameter perturbation. Adding oscillators improves spiral regularity while reducing resource capture, so the minimal sufficient circuit performs best.  We propose that this principle underlies search behavior in simple living systems, navigational and other decision-making in more complex ones, and, being so simple and ubiquitous, goes almost unnoticed unless you reduce the logic to a bare minimum. Eventually, networks of such proactive primitives may offer an alternative foundation for artificial intelligence architectures that explore our world rather than merely predict the next symbol in a sequence.
\end{abstract}

\section{Introduction}

Adaptive switching between broad exploration and focused exploitation is one of the most conserved behavioral motifs in motile organisms and one of the most studied trade-offs in organizational and cognitive science, in which search effort is adjusted to the strength of local evidence {\color{refblue}\cite{ref1}}. Its minimal known instantiation is striking: \textit{C. elegans} implements a multi-stage foraging strategy using a circuit of just a few neurons, switching from locally intensive search to wide-ranging exploration as the probability of finding food in the immediate area decays {\color{refblue}\cite{ref2}}. Subsequent work has dissected the parallel circuit mechanisms underlying this transition {\color{refblue}\cite{ref3}}, linked foraging decisions to multi-timescale coupling {\color{refblue}\cite{ref4}}, and offered a stochastic alternative to discrete state-driven switching {\color{refblue}\cite{ref5}}. This behavior is a specific instance of area-restricted search (ARS), defined by adaptive switching logic between directional, global exploration and focused, local exploitation in response to resource encounters {\color{refblue}\cite{ref6}}.

The simplicity of the \textit{C. elegans} circuit is not incidental: it suggests that the computational logic underlying ARS need not be elaborate, and can arise from a small number of interacting elements whose dynamics encode the transition. However these results do not isolate the minimal circuit sufficient to produce the switch.

At the same time, similar decentralized rules can produce globally efficient behavior in other biological systems. In \textit{P. polycephalum}, adaptive growth dynamics yield transport networks comparable in cost and fault tolerance to human-engineered infrastructure {\color{refblue}\cite{ref7}}, and the same system has been shown to solve NP-hard problems in linear-time {\color{refblue}\cite{ref8}}. Together with other oscillatory and decentralized biological examples, this suggests that efficient search and coordination can emerge from relatively simple local dynamics.

What remains unclear is how little structure is sufficient for such switching behavior to emerge. That question matters computationally as well as biologically. Reproducing this class of behavior in artificial agents requires a substrate sensitive to timing relationships between elements rather than one that operates on scalar activations alone. In artificial systems, however, spiking neural networks remain much harder to train than conventional ANNs, especially in the unsupervised regime, where optimization targets are less clear {\color{refblue}\cite{ref9,ref10}}. This difficulty matters here because recent surveys argue that temporal dynamics remain an underexploited aspect of current SNN design: many methods still import ANN-centered assumptions and therefore underuse spike timing and multi-timescale integration {\color{refblue}\cite{ref11}}. At the same time, compact spiking systems have already been shown to support klinokinetic resource search on neuromorphic hardware and adaptive path planning through axonal-delay mechanisms {\color{refblue}\cite{ref12,ref13}}. In parallel, insect-inspired spiking models have produced adaptive visual navigation and foraging-oriented sensorimotor control {\color{refblue}\cite{ref14,ref15}}. Reservoir approaches similarly exploit untrained recurrent dynamics for temporal computation, but rely on larger recurrent substrates plus a trained readout {\color{refblue}\cite{ref16}}. Oscillatory neural networks (ONNs) compute directly with coupled oscillators, encoding information in phase and frequency relationships and targeting pattern retrieval, image processing, and combinatorial optimization {\color{refblue}\cite{ref17}}. The present circuit differs in architecture and function. Its units are not coupled to one another; they share an input and sum onto the same motor channels. No coupling weights are learned or adapted at runtime, and the informative regime is the transient disagreement between fast and slow temporal evaluators, not a settled phase configuration. These systems show that adaptive search can be built and that oscillatory dynamics can compute, but they do not isolate how little structure is sufficient for it to emerge. The specific question driving this study is therefore narrower: what is the simplest oscillatory arrangement that produces autonomous exploration-exploitation switching, and what structural conditions are necessary for it to arise?

We address that question within the framework introduced in {\color{refblue}\cite{ref18}}, which proposed natural neural networks as an evolutionary-trained foundation for unsupervised artificial neural networks and identified a family of increasingly complex continuous-time kinematic agents, culminating in a composite spiking agent that exhibited outward spiral trajectories in sparse environments and sinusoidal tracking along structured food sources. That study established the behavioral repertoire qualitatively, but did not determine whether the switching was a robust emergent property of the circuit or an artifact of specific parameter choices.

This paper tests that point directly. We construct a composite circuit from frequency-tuned spiking oscillatory units (inverters) arranged in an asymmetric left-right topology with counter-phase wiring. Each inverter evaluates the same input over a different temporal window and inverts its response: it fires actively when input is absent and suppresses output when input is detected. The resulting behavior (spiral search in void, tracking near resources, and autonomous switching between them) is not explicitly programmed but emerges from the interaction of units operating at staggered timescales on the same input-output channel.

Through systematic hypothesis testing across 63 configurations (10 hypotheses plus 11 negative controls, 1,000 independent trials each, 95 of 95 validation checks passed), we demonstrate that: (i) the spiral is a genuine emergent property requiring three or more inverters, and is structural rather than noise-driven at zero rotational diffusion (H7 hypothesis); (ii) it depends on both temporal staggering and counter-phase opposition; neither condition alone suffices, and the positive control exceeds the best ablation by $3.1{\times}$ (H2); (iii) the circuit autonomously transitions from exploration to exploitation on encountering structured resources, consuming $3.2{\times}$ more food than single-inverter controls (H3); (iv) the behavior operates across a continuous range of opposition strengths and degrades gracefully under parameter perturbation (H4a--b); and (v) more inverters produce better spirals but catch less food (H6), a finding confirmed by exploratory follow-up H8 and implying that the most efficient foraging agent is the minimal circuit with sufficient spiral structure.

Our claim is correspondingly narrow. The composite oscillator is presented as a minimal synthetic mechanism for adaptive switching, not as a general definition of intelligence. Its core principle is temporal disagreement between concurrent evaluations of the same signal at different timescales: when short- and long-window evaluations diverge, the system searches; when they converge, it settles into a stable mode. This mechanism is directly testable in biological systems and may also provide a tractable building block for artificial systems operating under uncertainty.

\section{Core Definitions}

The following definitions are used throughout this paper. Quantitative metrics used to operationalize each concept are detailed in Section 4.2.

\textbf{Spiral trajectory.} A trajectory exhibiting self-similar curvature that expands outward over time, operationally classified by positive \textit{spiral growth rate} and \textit{spiral quality} exceeding a predefined threshold.

\textbf{Exploratory motion.} Movement characterized by outward spatial growth, low revisitation, and expanding coverage. Spiral trajectories in void environments are the prototypical instance.

\textbf{Exploitative motion.} Movement that remains bounded near a resource, with high revisitation and directional persistence along the resource gradient. Sinusoidal tracking along a food line is the prototypical instance.

\textbf{Emergent behavior.} A system-level behavior arising from the interaction of components under specific structural conditions. A behavior qualifies as emergent only if (i) no single component produces it in isolation, (ii) it appears reliably when structural conditions are met, and (iii) removing any of those conditions abolishes the behavior. This operational definition is tested through systematic ablation (H2).

\textbf{Exploration-exploitation switching.} The autonomous transition between exploratory and exploitative modes in response to environmental input, without an explicit controller or state machine managing the transition. This is the central phenomenon under investigation.

\textbf{Temporal disagreement.} The condition in which concurrent temporal evaluators, operating on the same input signal at different timescales, produce conflicting assessments of the current state. In a composite circuit, a fast evaluator may detect a signal change while a slow evaluator has not yet accumulated sufficient evidence to respond. We hypothesize that this disagreement is the causal driver of exploration-exploitation switching: when evaluators agree (all detect signal, or all detect absence), the system commits to a stable mode; when they disagree, the system is in transition between modes, producing the expanding spiral as a geometric consequence of partially converging motor commands.

\textbf{Signal and food.} The inverter circuit responds to any binary input signal (presence or absence of events within its decision window). In this study, we instantiate the input signal as a spatial food source to ground the simulations in a concrete foraging scenario.

\section{Model}

This section describes the composite spiking oscillator from first principles. We begin with the atomic unit (the inverter), then show how multiple inverters are wired into a composite, derive the motion dynamics mathematically, and explain the mechanism by which spiral search emerges.

\subsection{The Inverter}

The fundamental building block is the \textbf{inverter} (INV), a minimal spiking decision unit with one input channel and two output channels (Left and Right). The inverter operates by frequency inversion: it counts input signals over a temporal decision window and inverts its response.

\textbf{Operation.} The inverter accumulates input events over a window of duration $C_1$ seconds. When the window completes:

\begin{itemize}
\item If one or more input signals were detected during the window, the inverter enters the \textbf{suppressed state} (LOW output): it fires Left spikes at period $C_2$ and Right spikes at period $C_4$.
\item If no input signals were detected, the inverter enters the \textbf{active state} (HIGH output): it fires Left spikes at period $C_1$ and Right spikes at period $C_3$.

\end{itemize}
The window then restarts. The inverter remains silent until its first decision window completes; it must observe a full $C_1$ interval before making its first decision.

Formally, let N(t, t + $C_1$) denote the number of input events detected during a decision window of duration $C_1$ starting at time t. The output periods for the Left channel ($f_{o1}$) and Right channel ($f_{o2}$) are:

\begin{equation}
f_{o1} = \begin{cases} C_1, & N(t,\,t{+}C_1) = 0, \\ C_2, & N(t,\,t{+}C_1) \geq 1, \end{cases} \qquad
f_{o2} = \begin{cases} C_3, & N(t,\,t{+}C_1) = 0, \\ C_4, & N(t,\,t{+}C_1) \geq 1. \end{cases}
\end{equation}
When no input event arrives within the decision window (N = 0), the inverter fires at active-state periods ($C_1$, $C_3$). When one or more events are detected ($N \geq 1$), it fires at suppressed-state periods ($C_2$, $C_4$). The windowed event-count formulation used here directly matches the implementation.

\textbf{Parameters.} Each inverter is defined by four period parameters ({\color{refblue}\mbox{\textup{Table 1}}}):

\begin{table}[ht]
\centering
\small
\begin{tabular}{ll}
\toprule
\textbf{Parameter} & \textbf{Role} \\
\midrule
$C_1$ & Decision window length; Left output period in active (HIGH) state \\
$C_2$ & Left output period in suppressed (LOW) state \\
$C_3$ & Right output period in active (HIGH) state \\
$C_4$ & Right output period in suppressed (LOW) state \\
\bottomrule
\end{tabular}
\caption{Inverter period parameters}
\end{table}

\textbf{Frequency constraint.} The model requires an asymmetric relationship between Left and Right output frequencies that flips between states. In period space:

\begin{equation}
C_3 < C_1 < C_2 < C_4
\end{equation}
Equivalently, in frequency space: f($C_3$) > f($C_1$) > f($C_2$) > f($C_4$). This constraint ensures that in the active state (no input), Right fires faster than Left, producing a rightward curve. In the suppressed state (input detected), Left fires faster than Right, producing a leftward deflection. The asymmetry reversal between states is what enables the sinusoidal tracking pattern along food sources: the agent oscillates between leftward and rightward turns as it crosses and re-crosses the food boundary.

\textbf{Inter-spike variability.} Each spike interval is drawn from a Gaussian distribution centered on the nominal period with a coefficient of variation (CV) of 0.1, clamped to [$0.5{\times}$, $1.5{\times}$] the base period. This models biological inter-spike interval variability without affecting the mean rate.

\subsection{Composite Wiring}

A composite agent consists of $N \geq 3$ inverters that share the same input signal and whose outputs are summed to drive the same pair of motors (Left, Right). The first inverter ($f_1$) is wired \textbf{normally}: its Left output drives the Left motor and its Right output drives the Right motor. All subsequent inverters ($f_2$, $f_3$, ...) are wired in \textbf{counter-phase} (crossed): their Left output drives the Right motor and vice versa. {\color{refblue}\mbox{\textup{Figure 1}}} illustrates the general topology: blue arrows indicate normal wiring ($f_1$), red arrows indicate counter-phase wiring ($f_2$ through $f_N$).

\begin{figure}[ht]
\centering
\begin{tikzpicture}[
    node distance=1.4cm and 2.2cm,
    inv/.style={circle, draw, very thick, minimum size=0.9cm, font=\normalsize\itshape},
    invN/.style={circle, draw, very thick, minimum size=0.9cm, font=\normalsize\itshape, dashed},
    motor/.style={rectangle, draw, thick, minimum width=0.7cm, minimum height=0.5cm,
                  font=\small},
    normal/.style={->, very thick, >=stealth, color=blue!70!black},
    crossed/.style={->, thick, >=stealth, color=red!70!black},
    input/.style={->, very thick, >=stealth},
  ]
  \node[inv, blue!70!black] (f1) {$f_1$};
  \node[inv, red!70!black, below=of f1] (f2) {$f_2$};
  \node[below=0.6cm of f2, font=\Large, color=red!70!black] (dots) {$\vdots$};
  \node[invN, red!70!black, below=0.6cm of dots] (fN) {$f_N$};

  \node[motor, right=of f1, yshift=-1.4cm] (L) {L};
  \node[motor, right=of fN, yshift=+1.4cm] (R) {R};

  \node[motor, left=1.8cm of f2] (IN) {IN};
  \draw[input, color=blue!70!black] (IN) -- (f1);
  \draw[input, color=red!70!black] (IN) -- (f2);
  \draw[input, color=red!70!black, dashed] (IN) -- (fN);

  \draw[normal] (f1) -- (L);
  \draw[normal] (f1) -- (R);

  \draw[crossed] (f2) -- (L);
  \draw[crossed] (f2) -- (R);

  \draw[crossed, dashed] (fN) -- (L);
  \draw[crossed, dashed] (fN) -- (R);
\end{tikzpicture}
\caption{Composite wiring topology for an N${\times}$1${\times}$2 counter-phase circuit. Blue: normal wiring ($f_1$). Red: counter-phase wiring ($f_2$ through $f_N$).}
\end{figure}
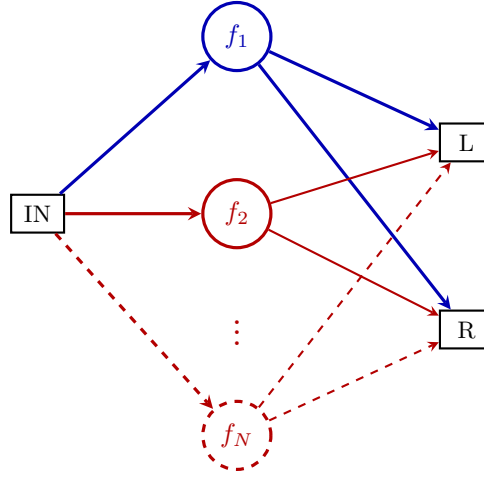

The key structural features are:

\begin{enumerate}
\item \textbf{Temporal staggering.} Each inverter has a different $C_1$ period: $f_1$ has the shortest (e.g., 0.15 s), $f_2$ is longer (e.g., 3.0 s), $f_3$ longer still (e.g., 6.0 s). After a reset, each inverter completes its decision window at a different time, producing a staggered activation sequence.

\item \textbf{Counter-phase opposition.} Crossed inverters oppose $f_1$'s turn differential. The opposition fraction decreases with each additional inverter (default: 27\%, 15\%, 5\% of $f_1$'s differential for $f_2$, $f_3$, $f_4$ respectively), preventing full reversal while progressively widening the turn radius.

\end{enumerate}
\textbf{Auto-recalculation.} Given $f_1$'s parameters and each crossed inverter's $C_1$, the remaining parameters ($C_2$, $C_3$, $C_4$) are derived automatically. For a crossed inverter with opposition fraction $\alpha$:

\begin{equation}
\frac{1}{C_{3,\text{crossed}}} = \alpha\left(\frac{1}{C_{3,f1}} - \frac{1}{C_{1,f1}}\right) + \frac{1}{C_{1,\text{crossed}}}.
\end{equation}
The $C_2$ and $C_4$ values are then set by fixed ratios, $C_2 = R_1 \cdot C_1$ and $C_4 = R_2 \cdot C_2$, with $R_1$ = 1.33 and $R_2$ = 2.0, to satisfy the NNN frequency constraint.

\textbf{Output combination.} At each timestep, all inverters receive the same input signal and produce independent spike outputs. The motor signals are computed by linear summation:

\begin{equation}
L_{\text{total}} = \sum_{\text{normal}} L_i + \sum_{\text{crossed}} R_i, \qquad
R_{\text{total}} = \sum_{\text{normal}} R_i + \sum_{\text{crossed}} L_i.
\end{equation}
\subsection{Motion Dynamics}

The combined motor signals drive a differential-drive kinematic agent through the following steps:

\textbf{Power.} Each inverter contributes its firing rate (Hz) directly to the motor channel, with no additional gain or baseline:

\begin{equation}
P = \frac{1}{\text{period}}
\end{equation}
The total Left and Right power values are the sums across all inverters (with crossed routing applied).

\textbf{Turn angle.} The heading change per timestep is a saturating function of the Left-Right power difference:

\begin{equation}
\delta\theta = 90^{\circ} \cdot \operatorname{sign}(\Delta P) \cdot \bigl(1 - e^{-k\,|\Delta P|}\bigr)
\end{equation}
where $\Delta P = P_{\mathrm{right}} - P_{\mathrm{left}}$ and $k = 0.08$ is the angular proportionality constant. The result is in degrees, converted to radians for the kinematic update. The saturation at $\pm$90 degrees prevents the agent from spinning in place.

\textbf{Speed.} Forward speed is proportional to total power:

\begin{equation}
v = s \cdot (P_{\text{right}} + P_{\text{left}})
\end{equation}
where $s = 5.0$ is the speed scaling factor. Speed is measured in pixels per second.

\textbf{Rotational diffusion.} Per-step Gaussian heading noise models biological motor variability. Let $\Delta t$ denote the simulation timestep (1/60 s at 60 fps):

\begin{equation}
\eta = \sqrt{2\,D_{\text{rot}}\,\Delta t}\;\xi,
\qquad \xi \sim \mathcal{N}(0,\,1)
\end{equation}
with $D_{\mathrm{rot}} = 0.1\;\mathrm{rad}^2/\mathrm{s}$. Because the amplitude scales with the square root of the timestep, the variance of the heading increment grows linearly with it, as a diffusion process requires. This follows the Active Brownian Particle model {\color{refblue}\cite{ref19}} and provides organic path variation without affecting the mean trajectory structure. The noise is added to the deterministic heading change at each timestep.

\textbf{Kinematic update.} At each timestep, the agent updates its heading by ($\delta\theta + \eta$), then moves forward by ($v \cdot \Delta t$) in the new heading direction.

\subsection{Spiral Emergence Mechanism}

The outward spiral is not programmed; it emerges from the staggered activation of inverters with different $C_1$ periods. When the input signal changes, all inverters clear their decision windows and re-enter the counting phase. Each must then complete a full $C_1$ interval before firing again. Because the $C_1$ values differ, the inverters reactivate at staggered times. Consider a composite of four inverters ($f_1$ normal, $f_2 - f_4$ crossed) in a void environment (no food):

\begin{enumerate}
\item \textbf{t = 0}: All inverters reset. No output from any unit.

\item \textbf{t = $C_{1,f1}$} (0.15 s): $f_1$ completes its decision window first. No input was detected, so $f_1$ enters the active state. Its Right fires faster than Left (1/$C_3$ > 1/$C_1$), producing a tight rightward circle. Effective differential: +0.74 Hz.

\item \textbf{t = $C_{1,f2}$} (3.0 s): $f_2$ completes its window. Being crossed, its output opposes $f_1$'s differential by 27\%, reducing the net differential to +0.54 Hz. The turn radius widens.

\item \textbf{t = $C_{1,f3}$} (6.0 s): $f_3$ activates, opposing by an additional 15\%. Net differential: +0.43 Hz. The radius widens further.

\item \textbf{t = $C_{1,f4}$} (10.0 s): $f_4$ activates, opposing by 5\%. Net differential: +0.39 Hz. Widest arc.

\end{enumerate}
The progressive widening of the turn radius combined with continuous forward motion produces an expanding spiral trajectory. Each stage is a discrete arc segment at a wider radius than the previous one. The spiral is self-similar: the ratio of successive radii is determined by the opposition fractions, which are fixed structural parameters.

\textbf{Mode switching and its causal mechanism.} When the agent encounters food, all inverters receive the input signal simultaneously. However, they do not respond simultaneously; each must complete its own $C_1$ window before making a decision. The fast inverter ($C_1$ = 0.15 s) responds within 150 milliseconds: it detects food and transitions to the suppressed state, producing a sharp leftward deflection toward the food source. Meanwhile, the slow inverters ($C_1$ = 3.0 s, 6.0 s) are still counting; their windows have not yet elapsed, so they continue to produce their void-state output (rightward bias). During this transitional period, the fast and slow inverters \textit{disagree}: the fast inverter says "goal is here, turn toward it," while the slow inverters still say "no goal detected, continue the arc." The result is a gradual narrowing of the spiral as each successive inverter reaches its decision threshold and joins the consensus.

If food contact persists, eventually all inverters complete their windows and converge on the suppressed state, producing the slow, sharp-turning tracking pattern characteristic of the single inverter. If food contact ceases, the staggered activation sequence restarts, and the spiral re-emerges. The transition between exploration (spiral) and exploitation (tracking) is thus driven by the progressive convergence or divergence of temporal evaluations at different timescales. When all windows agree (no food or sustained food), the circuit commits to a stable mode. When they disagree (transient signal change), the circuit is in transition. No explicit state machine or controller is required; the mode switch is a direct consequence of temporal disagreement between concurrent evaluators.

\textbf{Multi-scale temporal integration.} The staggered activation mechanism can be understood as an instance of a broader computational principle: evaluating the same signal through multiple concurrent temporal windows to obtain a multi-resolution representation of temporal structure. A short window ($C_{1,f1}$ = 0.15 s) responds rapidly to changes in input but cannot distinguish brief interruptions from sustained absence, whereas a long window ($C_{1,f3}$ = 6.0 s) integrates over extended periods, providing stability at the cost of responsiveness. Operating these windows in parallel effectively implements a temporal filter bank, in which fast components track rapid fluctuations and slow components track longer-term trends.

\section{Hypotheses and Methods}

\subsection{Experimental Framework}

All experiments are conducted in a virtual simulation environment. The agent operates on a 2D plane with configurable food geometries: void (empty plane), dense band (horizontal food band), and small circle (filled disk of food). The simulation runs at 60 frames per second with deterministic physics and stochastic rotational diffusion noise ($D_{\mathrm{rot}} = 0.1\;\mathrm{rad}^2/\mathrm{s}$). All angles in the motion model are computed in degrees and converted to radians for the kinematic update.

The simulation was implemented in Python 3.12 using Pygame 2.6 for rendering and NumPy 2.4 for numerical computation. Each trial used an independent random seed generated by Python's default OS-seeded random state (no fixed seeds). Simulation duration was 60 seconds per trial at 60 frames per second for all configurations except H1 and H7 (40 seconds). Each experimental configuration is run for 1,000 independent trials to establish statistical reliability. All metrics are computed from 1,000 independent trials. Key comparison metrics are reported with 95\% confidence intervals to characterize within-model variability; supporting metrics are reported as means. Comparisons between conditions use non-overlapping confidence intervals as the criterion for statistical separation. These intervals quantify the precision of estimates within the simulated environment and should not be interpreted as claims about biological systems.

To benchmark against standard foraging strategies from the ecology literature, we implemented five external baselines using the same physics pipeline: (i) a stochastic random walk (uniform random power at 1 Hz), (ii) a correlated random walk (CRW; persistence = 0.7, Gaussian heading perturbation), (iii) a Levy walk ($\mu = 2.0$, truncated power-law step durations), (iv) a prescribed logarithmic spiral ($r = a\,e^{b\theta}$, the theoretical ideal), and (v) an area-restricted search agent (ARS; rule-based switching from wide exploration to tight circling on food detection). Each baseline was tested in void, dense band, and dense bands environments for 1,000 trials (21,000 total baseline trials).

\subsection{Trajectory Metrics}

The following metrics are computed for each trial from the recorded (x, y) trajectory:

\textbf{Straightness.} The ratio of net displacement to total path length. A perfectly straight trajectory scores 1.0; a closed loop scores near 0.

\textbf{Circle fit score.} The $R^2$ goodness-of-fit of the trajectory to a least-squares circle. A perfect circle scores 1.0. The fitted radius is reported separately as \textit{circle fit radius}.

\textbf{Spiral quality.} An envelope-based composite metric defined as

\begin{equation}
Q = R^2(r,\,t) \cdot \left|\frac{1}{T} \sum_{k=1}^{T} e^{i\,\Delta\theta_k}\right|
\end{equation}
where r(t) = ||(x(t), y(t)) - centroid|| is the radial distance from the trajectory centroid, $R^2(r, t)$ is the coefficient of determination of an ordinary least-squares linear regression of r on t over the full trajectory (no binning or peak extraction), and the second factor is the mean resultant length of the per-frame heading increments $\Delta\theta_k$ (circular consistency in [0, 1]). Q is bounded in [0, 1]. For reference, on a textbook Archimedean spiral Q approaches 1.0; on a perfect circle $R^2(r, t)$ approaches 0 and Q approaches 0; on an isotropic random walk both factors are small and Q approaches 0. The metric is therefore high only when the trajectory expands outward monotonically \textit{and} maintains a consistent turning direction. A threshold sensitivity analysis confirms that the core findings (H1--H9) are stable across a range of spiral quality thresholds. Most checks pass for thresholds from 0.05 to 0.35; the narrowest per-check range is [0.05, 0.15] (H1a, 3-inverter spiral quality) and the most common stable range is [0.08, 0.35]. Individual per-check stability ranges are reported in Supplementary Table S1.

\textbf{Spiral growth rate.} The slope of the radial expansion envelope (pixels per frame). Positive values indicate outward growth; zero or negative values indicate stable or collapsing trajectories.

\textbf{Mean speed.} Average displacement per frame, converted to pixels per second.

\textbf{Area cells visited.} The number of distinct spatial grid cells ($5{\times}5$ pixel bins) entered during the trial. Measures the spatial extent of exploration.

\textbf{Revisitation rate.} The fraction of movement steps that enter a grid cell already visited. High revisitation indicates exploitative, locally bounded movement; low revisitation indicates broad exploration.

\subsection{Statistical Analysis}

\textbf{Estimands.} The primary estimand for each configuration is the population mean of the per-trial metric (spiral\_quality, food\_eaten, area\_cells\_visited, mean\_speed, revisitation\_rate, or circle\_fit\_score, depending on the hypothesis). All reported values are sample means over N = 1,000 independent trials.

\textbf{Sample size.} Each configuration was run for 1,000 independent trials. The choice of N = 1,000 is justified by a Monte Carlo standard-error argument: for the primary comparison metric (spiral\_quality in H2g, mean = 0.395, std = 0.292), the standard error at N = 1,000 is $\mathrm{SE} = 0.292/\sqrt{1000} = 0.009$, yielding a 95\% CI half-width of 0.018, less than 5\% of the mean. For food\_eaten in H3a (mean = 283.9, std = 62.4), SE = 1.97, giving a half-width of 3.87, or 1.4\% of the mean. A convergence analysis confirms that the standard error of the mean decreases as $1/\sqrt{N}$ across all key estimands, and that effect-size estimates stabilize well below N = 1,000 (Supplementary Figure S1).

\textbf{Confidence intervals.} 95\% confidence intervals are constructed using the normal approximation: $\mathrm{CI} = \mathrm{mean} \pm 1.96\,\mathrm{std}/\sqrt{N}$. With N = 1,000 and approximately symmetric distributions for most metrics, the normal approximation is appropriate by the central limit theorem. For metrics with substantial skewness (e.g., spiral\_quality in stochastic controls, skewness > 2.0), we verify that conclusions hold using median and interquartile range; distribution-level statistics including median, IQR, and skewness for all 84,000 trials are provided in the supplementary materials (Tables S2--S12, Figures S2--S12).

\textbf{Comparison criterion.} Comparisons between conditions use non-overlapping 95\% confidence intervals as the criterion for statistical separation. This is a conservative criterion: non-overlapping CIs at the 95\% level imply significance at approximately p < 0.006 for equal-variance comparisons, which is substantially more stringent than the conventional p < 0.05 threshold. For external baseline comparisons (Section 5.5), Cohen's d effect sizes with magnitude classifications (small: 0.2, medium: 0.5, large: 0.8, very large: 1.2) are reported alongside descriptive statistics.

\textbf{Multiplicity.} The study evaluates 95 validation checks across 10 hypotheses and 11 negative controls. We address multiplicity through three complementary strategies rather than a single correction factor: (i) the non-overlapping CI criterion is itself conservative (effective $\alpha$ approximately 0.006 per comparison); (ii) the negative-control battery (H9) provides 13 independent checks that the metrics discriminate genuine structure from noise (if the metrics were spuriously detecting patterns, negative controls would pass rather than fail); and (iii) a threshold sensitivity analysis (Supplementary Table S1) sweeps the primary decision thresholds across the range 0.05--0.35, confirming that all core findings are stable across a range of thresholds (most common range [0.08, 0.35]; narrowest per-check range [0.05, 0.15]; see Supplementary Table S1 for individual ranges). No individual check depends on a single threshold value. Formal test statistics (Welch's t-test for comparisons, one-sample t-test for thresholds, TOST for tolerance checks, bootstrap for ratio checks) with Holm-Bonferroni corrected p-values, Cohen's d effect sizes, and Mann-Whitney U nonparametric robustness checks for all 95 validation checks are reported in Supplementary Table S14. All directional tests are one-tailed, consistent with the prespecified direction of each check. The table reports two distinct columns: "Observed" (directional pattern holds in sample means) and "Inferred" (statistically supported after Holm-Bonferroni correction at adj p < 0.05). Of the 95 checks, 84 are confirmatory (H0--H5, H7, H9) and 11 are exploratory (H6: 6, H8: 5); H6 and H8 were introduced after observing unexpected results and are reported as exploratory analyses throughout. All 95 checks are observed in the predicted direction; 92 are inferentially supported; the 3 non-inferred checks are in H6 (exploratory) where adjacent configurations differ by d < 0.10. Key test statistics for the principal comparisons are reported inline in Section 5.

\textbf{Assumptions.} Trials are independent (each uses an OS-seeded random state with no fixed seeds). The simulation physics is deterministic conditional on the random seed; stochasticity enters only through rotational diffusion noise and inter-spike interval jitter. Metrics are computed from full trajectory recordings with no post-hoc selection of trials or time windows.

\subsection{Hypotheses}

We test ten hypotheses plus a comprehensive negative-control battery. Below we present four showcase hypotheses in full detail; these carry the core argumentative weight of the paper. The remaining six supporting hypotheses and the negative-control battery are summarized in compact form; full results for all are reported in Section 5.

\subsubsection{H2: Specificity (Ablation)}

\textbf{Prediction.} The spiral requires \textit{both} counter-phase opposition (crossed wiring) \textit{and} temporal staggering (different $C_1$ periods). Removing either condition abolishes the spiral. This is the strongest test of genuine emergence: we identify the necessary structural conditions and show that each is individually insufficient.

\textbf{Configurations (7).} All in void environment ({\color{refblue}\mbox{\textup{Table 2}}}):

\begin{table}[ht]
\centering
\small
\begin{tabular}{lllll}
\toprule
\textbf{ID} & \textbf{Config} & \textbf{Opposition} & \textbf{Staggering} & \textbf{Prediction} \\
\midrule
H2a & Stochastic agent & -- & -- & Random walk, no spiral \\
H2b & 1 inverter & -- & -- & Stable circle \\
H2c & 2-inv (N + C) & Yes & Yes & Wider arc, no spiral (insufficient inverters) \\
H2d & 2-inv (N + N) & No & Yes & Faster circle \\
H2e & 3-inv, all same $C_1$ & Yes & \textbf{No} & Reversed circle (opposition without staggering) \\
H2f & 3-inv, all normal & \textbf{No} & Yes & Accelerating circle (staggering without opposition) \\
H2g & 3-inv, correct wiring & Yes & Yes & Outward spiral (positive control = H1a) \\
\bottomrule
\end{tabular}
\caption{H2 ablation configurations}
\end{table}

\textbf{Key checks (11).} H2a--H2f: spiral\_quality < 0.20. H2g: spiral\_quality > 0.20 and > $2{\times}$ max(H2a--H2f). Phase transition evidence: H2g >> H2e and H2g >> H2f. H2g area > H2b area.

\subsubsection{H3: Exploitation (Mode Switching)}

\textbf{Prediction.} A composite agent spiraling in void autonomously transitions to exploitative tracking on encountering a structured food source. The transition is input-driven with no explicit controller. Single- inverter controls track food but do not exhibit the prior exploratory phase.

\textbf{Configurations (4).} See {\color{refblue}\mbox{\textup{Table 3}}}.

\begin{table}[ht]
\centering
\small
\begin{tabular}{llll}
\toprule
\textbf{ID} & \textbf{Agent} & \textbf{Environment} & \textbf{Prediction} \\
\midrule
H3a & Composite (4-inv) & Dense band & Spiral $\to$ tracking transition \\
H3b & Composite (4-inv) & Small circle & Spiral $\to$ tracking transition \\
H3c & Single inverter & Dense band & Tracking only (control) \\
H3d & Single inverter & Small circle & Tracking only (control) \\
\bottomrule
\end{tabular}
\caption{H3 mode-switching configurations}
\end{table}

\textbf{Key checks.} H3a/b: spiral\_quality > 0.15 in the pre-contact phase (lower than H2's 0.20 threshold because pre-contact spiral quality is measured over a shorter observation window before the agent encounters food); revisitation\_rate increases after contact; area coverage growth rate decreases after contact. H3c/d: no spiral phase detected; tracking begins immediately on food encounter. H3a/b revisitation\_rate (post-contact) > H3c/d revisitation\_rate.

\subsubsection{H6: Complexity Scaling}

\textbf{Prediction.} At fixed net opposition (Net $\sim$ +0.22), distributing opposition across more crossed inverters (3 to 7) improves food search efficiency.

\textbf{Outcome.} Contrary to our initial hypothesis we found that food efficiency decreased monotonically with inverter count. We report this as an exploratory finding and test its relationship to spiral regularity in H8.

\textbf{Configurations (5).} All composites at Net $\sim$ +0.22 in dense-band environment ({\color{refblue}\mbox{\textup{Table 4}}}):

\begin{table}[ht]
\centering
\small
\begin{tabular}{llll}
\toprule
\textbf{ID} & \textbf{Inverters} & \textbf{$C_1$ Values (s)} & \textbf{Prediction} \\
\midrule
H6a1 & 3 & 0.15, 0.75, 1.5 & Most food \\
H6a2 & 4 & 0.15, 0.5, 1.0, 2.0 & Slightly less food \\
H6a3 & 5 & 0.15, 0.5, 1.0, 2.0, 4.0 & Less food \\
H6a4 & 6 & 0.15, 0.5, 1.0, 2.0, 4.0, 8.0 & Less food still \\
H6a5 & 7 & 0.15, 0.5, 1.0, 2.0, 4.0, 8.0, 16.0 & Least food \\
\bottomrule
\end{tabular}
\caption{H6 complexity-scaling configurations}
\end{table}

\textbf{Key checks (6).} Food consumed is monotonically non-increasing from H6a1 to H6a5. H6a1 food > $1.3{\times}$ H6a5 food. All pairwise comparisons hold. Area visited follows the same monotonic ordering.

\subsubsection{H8: Reconciliation}

H8 was introduced after observing the H6 outcome and is reported as an exploratory follow-up analysis.

\textbf{Prediction.} The same configurations that produced decreasing food efficiency in H6 should produce \textit{increasing} spiral quality in void, confirming that regularity and efficiency are governed by the same parameters in opposite directions.

\textbf{Configurations (5).} Same as H6, but in void environment ({\color{refblue}\mbox{\textup{Table 5}}}):

\begin{table}[ht]
\centering
\small
\begin{tabular}{llll}
\toprule
\textbf{ID} & \textbf{Inverters} & \textbf{$C_1$ Values (s)} & \textbf{Prediction} \\
\midrule
H8a1 & 3 & 0.15, 0.75, 1.5 & Lowest spiral quality \\
H8a2 & 4 & 0.15, 0.5, 1.0, 2.0 & Higher quality \\
H8a3 & 5 & 0.15, 0.5, 1.0, 2.0, 4.0 & Higher quality \\
H8a4 & 6 & 0.15, 0.5, 1.0, 2.0, 4.0, 8.0 & Higher quality \\
H8a5 & 7 & 0.15, 0.5, 1.0, 2.0, 4.0, 8.0, 16.0 & Highest quality \\
\bottomrule
\end{tabular}
\caption{H8 reconciliation configurations}
\end{table}

\textbf{Key checks (5).} H8a5 spiral\_quality > H8a1. H8a5 > $1.3{\times}$ H8a1. H8a5 detection\_rate > 0.80. Detection rate increases with inverter count. Even the weakest configuration (H8a1) maintains spiral\_quality > 0.05.

\subsubsection{Supporting Hypotheses}

\textbf{H0: Baseline Dynamics (7 configs, 14 checks).} A single inverter with equal L/R frequencies produces straight-line motion; the NNN-constrained asymmetry produces stable circles. Speed scales linearly with absolute frequency while radius remains constant; widening the frequency ratio tightens the circle.

\textbf{H1: Emergence of Spiral Search (3 configs, 8 checks).} Composite agents with 3, 4, and 5 inverters ($f_1$ normal, $f_2$+ crossed) in void produce outward spiral trajectories. Spiral quality is non-decreasing with inverter count.

\textbf{H4a: Opposition Balance (7 configs, 12 checks).} The net opposition strength controls the trajectory character on a continuous gradient: strong positive differential produces a tight spiral, near-zero balance produces fast sweeping with maximal area coverage, and negative differential reverses the spiral direction.

\textbf{H4b: Robustness (4 configs, 5 checks).} Random jitter of $\pm$10\% to $\pm$30\% applied to all $C_1$ values degrades the spiral gracefully rather than catastrophically, supporting evolutionary plausibility.

\textbf{H5: Exploitation Geometry (4 configs, 7 checks).} Among composites with different net opposition strengths in a food environment, the most open spiral catches the most food. Food consumption increases monotonically as the spiral opens.

\textbf{H7: Noise Sensitivity (6 configs, 5 checks).} At $D_{\mathrm{rot}}$ = 0 (rotational diffusion removed, spike-timing jitter retained), the composite still produces a clear spiral with high quality and 100\% detection rate, proving the spiral is structural, not a noise artifact. Increasing noise degrades quality but increases area coverage.

\textbf{H9: Negative Controls (11 configs, 13 checks).} Eleven adversarial configurations, each designed to remove a specific structural feature tested by H0--H8, must fail the corresponding positive-hypothesis threshold, confirming the metric battery's discriminative power.

\subsection{Summary}

The full experimental design is summarized in {\color{refblue}\mbox{\textup{Table 6}}}.

\begin{table}[ht]
\centering
\small
\begin{tabular}{llll}
\toprule
\textbf{Hypothesis} & \textbf{Configs} & \textbf{Trials/Config} & \textbf{Total Trials} \\
\midrule
H0: Baseline & 7 & 1,000 & 7,000 \\
H1: Emergence & 3 & 1,000 & 3,000 \\
H2: Specificity & 7 & 1,000 & 7,000 \\
H3: Exploitation & 4 & 1,000 & 4,000 \\
H4a: Opposition Balance & 7 & 1,000 & 7,000 \\
H4b: Robustness & 4 & 1,000 & 4,000 \\
H5: Exploitation Geometry & 4 & 1,000 & 4,000 \\
H6: Complexity Scaling & 5 & 1,000 & 5,000 \\
H7: Noise Sensitivity & 6 & 1,000 & 6,000 \\
H8: Reconciliation & 5 & 1,000 & 5,000 \\
H9: Negative Controls & 11 & 1,000 & 11,000 \\
\textbf{Hypothesis subtotal} & \textbf{63} &  & \textbf{63,000} \\
External baselines (Section 5.5) & 21 & 1,000 & 21,000 \\
\textbf{Grand total (archived dataset)} & \textbf{84} &  & \textbf{84,000} \\
\bottomrule
\end{tabular}
\caption{Experimental design summary}
\end{table}

The 63 configurations above constitute the hypothesis-testing experiments (H0--H9 and the negative-control battery). The 21 external-baseline configurations introduced in Section 5.5 (five standard foraging strategies plus a stochastic control, each evaluated across multiple environments) add a further 21,000 trials, bringing the complete archived dataset to 84 configurations and 84,000 trials.

\section{Results}

All 95 validation checks passed across 63 configurations (1,000 trials each). Values are means; 95\% CIs are shown for key comparison metrics. Supporting-metric CIs are narrow (typically <3\% of mean) and are available in the supplementary materials (Tables S2-S12).

\subsection{Specificity (Ablation)}

This is the strongest test of genuine emergence. Six ablation conditions systematically remove one or both structural requirements (counter-phase opposition and temporal staggering). Only the positive control retaining both conditions produces a spiral ({\color{refblue}\mbox{\textup{Table 7}}}).

\begin{table}[ht]
\centering
\small
\begin{tabular}{lllll}
\toprule
\textbf{Config} & \textbf{Opposition} & \textbf{Staggering} & \textbf{Spiral Quality [95\% CI]} & \textbf{Area Visited} \\
\midrule
H2a (stochastic) & -- & -- & 0.038 [0.032, 0.044] & 38.3 \\
H2b (1 inverter) & -- & -- & 0.067 [0.062, 0.073] & 21.3 \\
H2c (2-inv N+C) & Yes & Yes & 0.127 [0.118, 0.137] & 49.4 \\
H2d (2-inv N+N) & No & Yes & 0.060 [0.055, 0.065] & 16.1 \\
H2e (same $C_1$) & Yes & \textbf{No} & 0.073 [0.067, 0.079] & 120.0 \\
H2f (all normal) & \textbf{No} & Yes & 0.088 [0.082, 0.095] & 15.0 \\
H2g (correct) & Yes & Yes & \textbf{0.395} [0.377, 0.413] & 82.6 \\
\bottomrule
\end{tabular}
\caption{H2 ablation results}
\end{table}

The positive control (H2g) achieves spiral quality $3.1{\times}$ the best ablation condition (H2c, 0.127; Welch's t = 30.2, p < 1e-50, d = 1.35). All six ablation conditions fall below the 0.20 threshold; the positive control exceeds it by nearly $2{\times}$ (one-sample t = 21.1, p < 1e-50, d = 0.67). Opposition alone (H2e, quality 0.073) and staggering alone (H2f, quality 0.088) each fail individually; both conditions are necessary, neither is sufficient. All 11 checks passed; all remain significant after Holm-Bonferroni correction (Supplementary Table S14). The results are summarized in {\color{refblue}\mbox{\textup{Figure 2}}}.

\begin{figure}[H]
\centering
\includegraphics[width=0.8\textwidth]{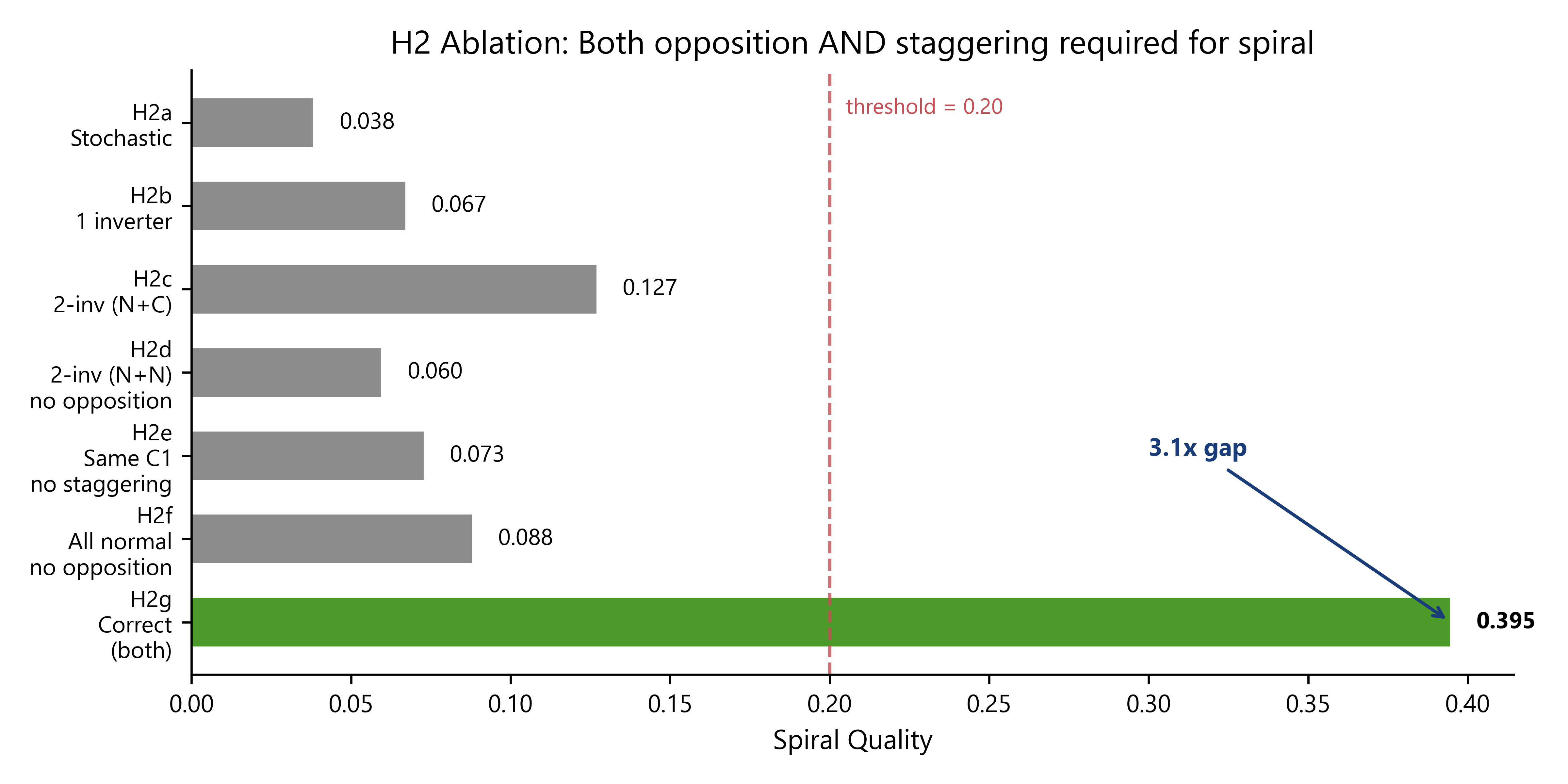}
\caption{H2 ablation results. Mean spiral quality across N = 1,000 trials per configuration. All six conditions lacking either opposition or staggering fall below the 0.20 spiral quality threshold. The positive control (H2g, both present) exceeds the best ablation by $3.1{\times}$.}
\end{figure}

\subsection{Mode Switching}

Composite agents spiral in void, then autonomously switch to exploitative tracking on encountering food. Single-inverter controls track food but never exhibit the prior exploratory phase ({\color{refblue}\mbox{\textup{Table 8}}}).

\begin{table}[ht]
\centering
\small
\begin{tabular}{llllll}
\toprule
\textbf{Config} & \textbf{Agent} & \textbf{Environment} & \textbf{Pre-food Spiral [95\% CI]} & \textbf{Food Eaten} & \textbf{Revisitation} \\
\midrule
H3a & Composite & Dense band & 0.413 [0.397, 0.430] & 283.9 & 0.964 \\
H3b & Composite & Small circle & 0.472 [0.456, 0.487] & 144.6 & 0.962 \\
H3c & Inverter & Dense band & 0.056 [0.045, 0.067] & 88.2 & 0.945 \\
H3d & Inverter & Small circle & 0.069 [0.057, 0.080] & 48.3 & 0.946 \\
\bottomrule
\end{tabular}
\caption{H3 mode-switching results}
\end{table}

The composite's pre-food spiral quality (0.41--0.47) is $7.3{\times}$ that of the inverter control (0.06--0.07), confirming a genuine exploratory phase. Composites consume $3.2{\times}$ more food on dense band and $3.0{\times}$ more on small circle (Cohen's d = 3.51 and 1.84). Post-contact revisitation exceeds 0.96, indicating complete transition to exploitation. The single inverter tracks effectively but lacks the prior search phase, making its success dependent on initial proximity. Full test statistics: Supplementary Table S14. A representative trajectory is shown in {\color{refblue}\mbox{\textup{Figure 3}}}.

\begin{figure}[ht]
\centering
\includegraphics[width=0.8\textwidth]{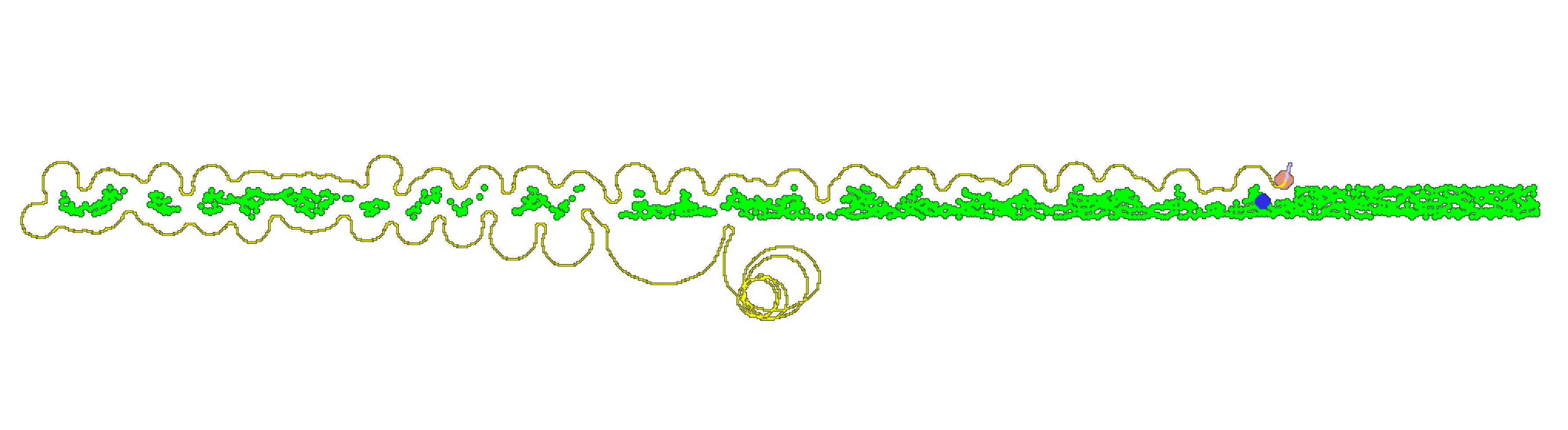}
\caption{H3 mode switching. Representative single-trial trajectory (selected as the trial closest to the median food\_eaten). A composite agent spawns in void (center), develops an outward spiral, encounters a dense food band, and transitions autonomously to sinusoidal tracking along the food source.}
\end{figure}

\subsection{Quality-Efficiency Dissociation}

\subsubsection{H6: Complexity Scaling}

The original prediction (that more inverters would improve foraging efficiency) was not confirmed. At fixed net opposition (Net $\sim$ +0.22), increasing the number of inverters from 3 to 7 monotonically \textit{decreases} efficiency ({\color{refblue}\mbox{\textup{Table 9}}}).

\begin{table}[ht]
\centering
\small
\begin{tabular}{llllll}
\toprule
\textbf{Config} & \textbf{Inverters} & \textbf{Food Eaten [95\% CI]} & \textbf{Area Visited} & \textbf{Mean Speed (px/s)} & \textbf{Revisitation} \\
\midrule
H6a1 & 3 & 749.0 [737.0, 760.9] & 165.2 & 181.5 & 0.909 \\
H6a2 & 4 & 733.8 [722.7, 744.9] & 165.1 & 209.3 & 0.909 \\
H6a3 & 5 & 654.7 [646.0, 663.5] & 140.5 & 208.3 & 0.923 \\
H6a4 & 6 & 593.1 [585.4, 600.9] & 117.6 & 198.0 & 0.935 \\
H6a5 & 7 & 530.2 [522.2, 538.3] & 96.6 & 193.5 & 0.947 \\
\bottomrule
\end{tabular}
\caption{H6 complexity-scaling results}
\end{table}

The 3-inverter composite eats 41.3\% more food than the 7-inverter (749.0 vs 530.2). Area coverage and revisitation follow the same monotonic trend. The mechanism: high-$C_1$ inverters in larger configurations take seconds to activate, diluting contributions and delaying the explore-to-exploit transition. All 6 checks confirmed. This motivated the H8 reconciliation analysis.

\subsubsection{H8: Reconciliation [Exploratory]}

The same inverter-count sweep that produced decreasing efficiency in H6 produces \textit{increasing} spiral quality in void, resolving the apparent contradiction ({\color{refblue}\mbox{\textup{Table 10}}}).

\begin{table}[ht]
\centering
\small
\begin{tabular}{llllll}
\toprule
\textbf{Config} & \textbf{Inverters} & \textbf{Spiral Quality} & \textbf{Detection Rate} & \textbf{Area Visited} & \textbf{Mean Speed (px/s)} \\
\midrule
H8a1 & 3 & 0.270 [0.255, 0.286] & 0.650 & 164.5 & 268.0 \\
H8a2 & 4 & 0.310 [0.293, 0.327] & 0.679 & 190.5 & 325.0 \\
H8a3 & 5 & 0.383 [0.366, 0.400] & 0.778 & 188.7 & 326.1 \\
H8a4 & 6 & 0.538 [0.522, 0.555] & 0.934 & 177.1 & 303.3 \\
H8a5 & 7 & \textbf{0.720} [0.709, 0.730] & \textbf{0.998} & 146.0 & 277.6 \\
\bottomrule
\end{tabular}
\caption{H8 reconciliation results}
\end{table}

Spiral quality increases $2.66{\times}$ from 3 to 7 inverters (0.270 to 0.720), while detection rate rises from 65.0\% to 99.8\%. The regularity comes at the cost of reduced spatial coverage: in the food environment of H6, the 3-inverter composite's wider sweeps encounter more food than the 7-inverter's tighter spiral. The optimal foraging agent is the minimal circuit with sufficient spiral structure. All 5 checks passed. The opposing trends are shown in {\color{refblue}\mbox{\textup{Figure 4}}}.

\begin{figure}[H]
\centering
\includegraphics[width=0.8\textwidth]{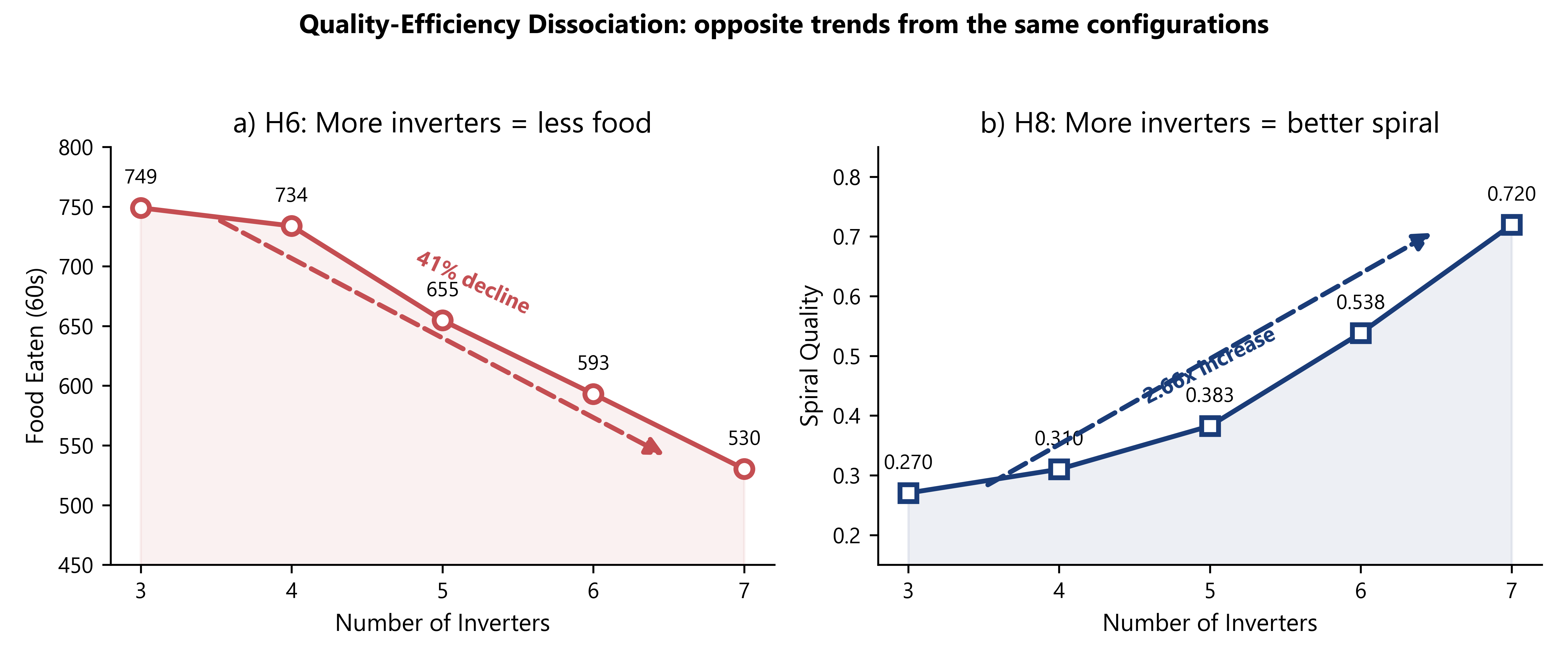}
\caption{Quality-efficiency dissociation. Mean values across N = 1,000 trials per configuration. (a) H6: more inverters catch less food in a foraging environment, declining 41\% from 3 to 7 inverters. (b) H8: the same configurations produce better spirals in void, increasing $2.66{\times}$. The optimal foraging agent is the minimal circuit.}
\end{figure}

\subsection{Supporting Results}

\subsubsection{H0: Baseline Dynamics}

A single inverter with equal L/R frequencies (H0a) produces near-straight motion (straightness = 0.097), while the NNN-constrained asymmetry produces stable circles. Speed scaling is precisely linear: H0d runs at $2.00{\times}$ H0b speed and H0e at $4.01{\times}$, while radius remains constant. Widening the frequency ratio tightens the circle monotonically (31.8 $\to$ 9.3 $\to$ 6.3 px). All 14 checks passed ({\color{refblue}\mbox{\textup{Fig. 5}}}).

\begin{figure}[H]
\centering
\includegraphics[width=0.8\textwidth]{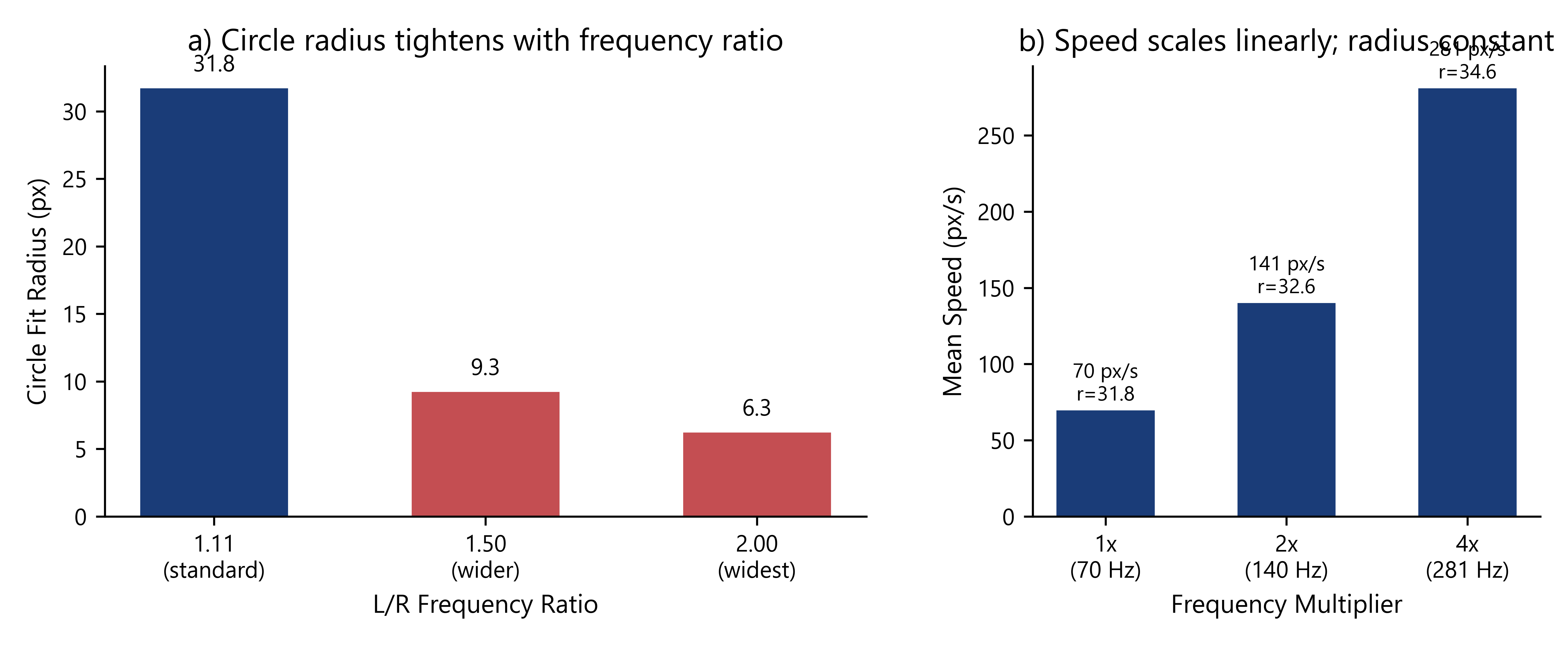}
\caption{Baseline single-inverter dynamics. Mean values across N = 1,000 trials per configuration. (a) Circle radius tightens monotonically with L/R frequency ratio. (b) Speed scales linearly with absolute frequency; radius remains constant.}
\end{figure}

\subsubsection{H1: Emergence of Spiral Search}

Composite agents with 3, 4, and 5 inverters all produce outward spirals in void. Spiral quality increases 76\% from the 3-inverter to the 5-inverter composite (0.165 to 0.290). All 8 checks passed.

\subsubsection{H4a: Opposition Balance}

The net opposition strength sweeps from +1.89 (tight forward spiral) through 0.0 (perfect balance) to -1.11 (reversed spiral). Area coverage increases $12.7{\times}$ from the tightest opposition to perfect balance. Peak spiral quality occurs at near-balance (Net = +0.22, quality 0.275), not at perfect balance, where the agent sweeps broadly but without spiral structure. All 12 checks passed.

\subsubsection{H4b: Robustness}

Applying per-trial uniform random jitter to all $C_1$ values degrades the spiral monotonically but never catastrophically. At 10\% jitter, 59.8\% of baseline quality is retained. Even at 30\% jitter, spiral quality (0.091) remains above the 0.05 detection threshold. All 5 checks passed.

\subsubsection{H5: Exploitation Geometry}

Among composites with different net opposition strengths, the most open spiral catches the most food. The progression is strictly monotonic across all four openness levels ({\color{refblue}\mbox{\textup{Table 11}}}).

\begin{table}[ht]
\centering
\small
\begin{tabular}{llllll}
\toprule
\textbf{Config} & \textbf{Net Diff} & \textbf{Food Eaten [95\% CI]} & \textbf{Area Visited} & \textbf{Mean Speed (px/s)} & \textbf{Revisitation} \\
\midrule
H5a1 (tight) & +1.89 & 456.3 [448.4, 464.2] & 48.9 & 156.9 & 0.973 \\
H5a2 (medium) & +0.89 & 559.6 [552.6, 566.5] & 78.3 & 150.3 & 0.957 \\
H5a3 (wide) & +0.45 & 636.5 [628.2, 644.8] & 128.3 & 147.7 & 0.929 \\
H5a4 (open) & +0.22 & 746.6 [734.8, 758.4] & 165.2 & 181.2 & 0.909 \\
\bottomrule
\end{tabular}
\caption{H5 exploitation geometry results}
\end{table}

The open spiral (H5a4) consumes 63.6\% more food than the tight spiral (H5a1). Area coverage expands $3.4{\times}$ from tight to open; revisitation decreases accordingly. All 7 checks passed.

The tight spiral (H5a1, Net = +1.89) confines the agent to a single food band ({\color{refblue}\mbox{\textup{Fig. 6a}}}). The open spiral (H5a4, Net = +0.22) sweeps wide enough to discover and exploit both bands ({\color{refblue}\mbox{\textup{Fig. 6b}}}), direct evidence that spiral geometry determines the agent's capacity to detect second-degree environmental symmetry. Performance traces are analyzed in Section 6.1.

\begin{figure}[H]
\centering
\begin{subfigure}[t]{0.48\textwidth}
\centering
\includegraphics[width=\textwidth]{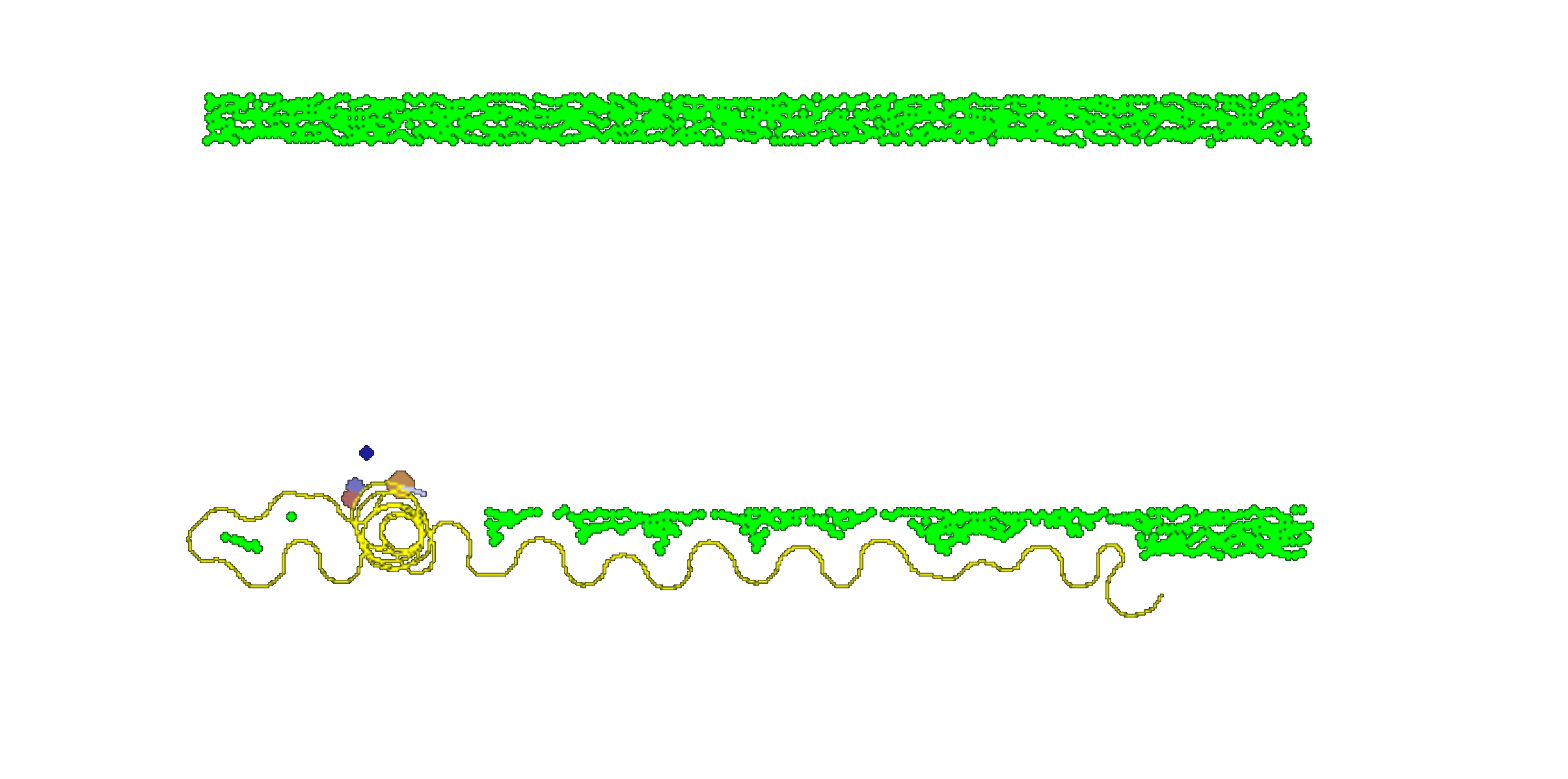}
\caption{H5 tight spiral (Net = +1.89): the agent tracks along one food band but never reaches the second.}
\end{subfigure}
\hfill
\begin{subfigure}[t]{0.48\textwidth}
\centering
\includegraphics[width=\textwidth]{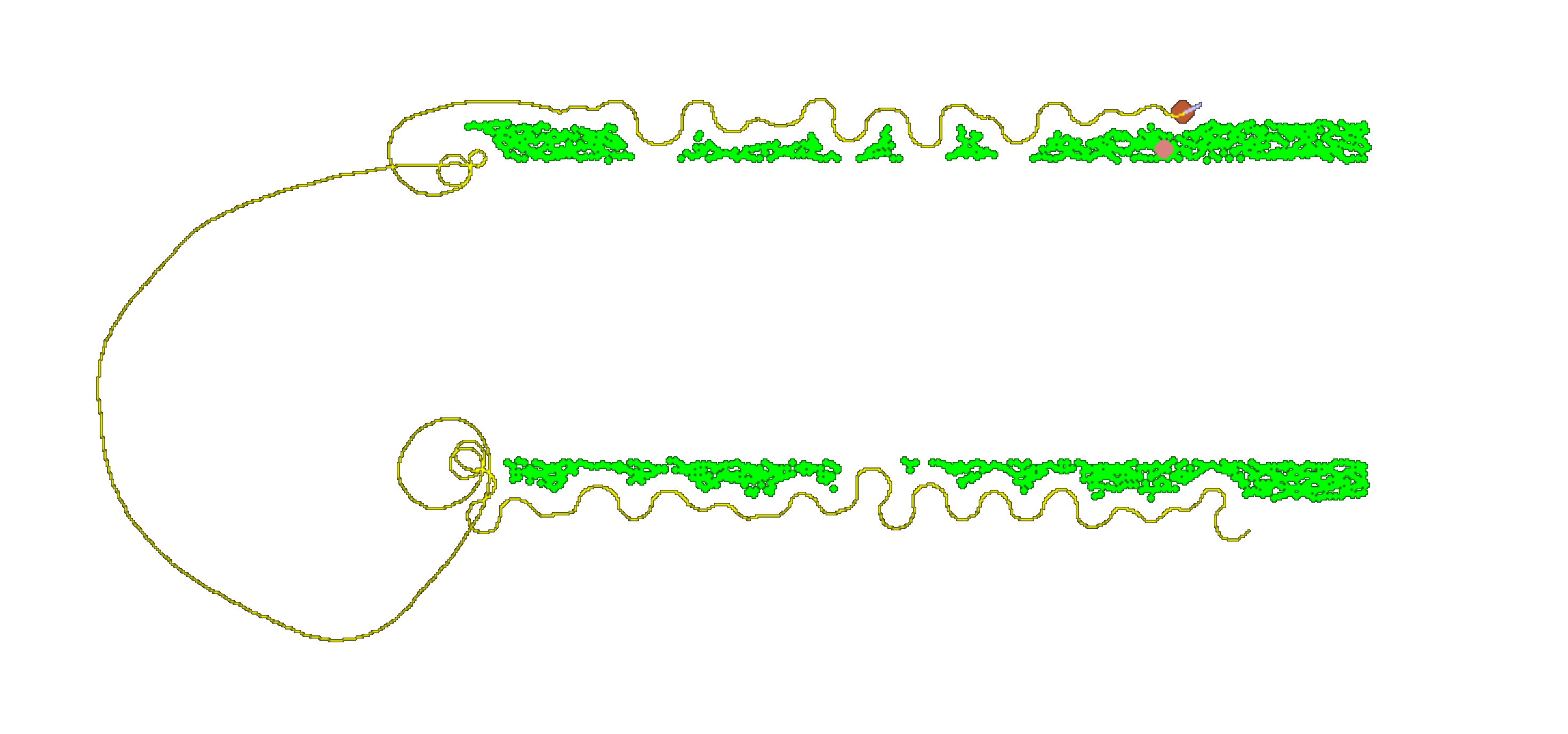}
\caption{H5 open spiral (Net = +0.22): the agent's wide arc reaches both food bands.}
\end{subfigure}
\caption{H5 exploitation geometry: tight vs.\ open spiral trajectories in the dense-bands environment.}
\end{figure}

\subsubsection{H7: Noise Sensitivity}

The zero-diffusion configuration ($D_{\mathrm{rot}}$ = 0) provides the strongest structural proof: with rotational diffusion removed, the composite produces a spiral with quality 0.652 and 100\% detection rate. Adding noise decreases quality (0.652 to 0.238) but increases area coverage (63.3 to 123.8 cells). The spiral remains detectable at all noise levels tested. All 5 checks passed.

\subsubsection{H9: Negative Controls}

Eleven adversarial configurations confirm that the metric battery discriminates genuine structure from noise and degenerate configurations. The single inverter at $D_{\mathrm{rot}}$ = 0 (neg\_h7) produces a perfect circle (circle\_fit\_score = 1.000) with spiral quality 0.004. The stochastic agent (neg\_h3) eats only 52.9 items (vs 283.9 for the composite in H3a). All 13 checks passed.

\subsection{External Baseline Comparison}

{\color{refblue}\mbox{\textup{Table 12}}} compares the 3-inverter NNN composite against five standard foraging baselines and a stochastic control, each run for 1,000 trials in the dense-bands environment.

\begin{table}[ht]
\centering
\small
\begin{tabular}{lllll}
\toprule
\textbf{Agent} & \textbf{Food Eaten [mean $\pm$ std]} & \textbf{Area Visited} & \textbf{Spiral Quality} & \textbf{Speed (px/s)} \\
\midrule
NNN (3-inv) & \textbf{483 $\pm$ 76} & 71 & 0.07 & 62 \\
NNN (7-inv) & \textbf{385 $\pm$ 131} & 51 & 0.13 & 83 \\
Log spiral & 228 $\pm$ 32 & 131 & 0.46 & 97 \\
Levy walk & 127 $\pm$ 98 & 111 & 0.06 & 126 \\
ARS & 126 $\pm$ 73 & 81 & 0.18 & 99 \\
CRW & 105 $\pm$ 79 & 114 & 0.06 & 131 \\
Stochastic & 61 $\pm$ 61 & 57 & 0.07 & 63 \\
\bottomrule
\end{tabular}
\caption{External baseline comparison (dense-bands, 1,000 trials each)}
\end{table}

The 3-inverter NNN composite consumes $2.1{\times}$ more food than the best external baseline (log spiral, Cohen's d = 4.35) and $7.9{\times}$ more than the stochastic control (d = 6.09). All food-eaten effect sizes exceed d = 4.0 (very large). Full effect-size tables are provided in Supplementary Table S13. The NNN agent achieves this with \textit{less} area coverage (71 vs 111--131 cells) and \textit{lower} speed (62 vs 97--131 px/s) than every baseline except stochastic. The advantage is not wider exploration but autonomous mode switching: the NNN agent slows and concentrates near food, while baselines traverse food regions without exploiting them.

In void ({\color{refblue}\mbox{\textup{Table 13}}}), the 7-inverter NNN composite produces spiral quality 0.67, compared to 0.07 for CRW, Levy, and ARS, and 0.33 for the prescribed log spiral. The NNN agents cover 150--167 cells vs 57 (stochastic) to 131 (log spiral).

\begin{table}[ht]
\centering
\small
\begin{tabular}{llll}
\toprule
\textbf{Agent} & \textbf{Spiral Quality [mean $\pm$ std]} & \textbf{Area Visited} & \textbf{Speed (px/s)} \\
\midrule
NNN (7-inv) & \textbf{0.67 $\pm$ 0.20} & 150 & 138 \\
Log spiral & 0.33 $\pm$ 0.42 & 131 & 89 \\
NNN (3-inv) & 0.25 $\pm$ 0.24 & 167 & 133 \\
Stochastic & 0.08 $\pm$ 0.11 & 57 & 61 \\
CRW & 0.07 $\pm$ 0.09 & 114 & 128 \\
Levy & 0.07 $\pm$ 0.09 & 110 & 124 \\
ARS & 0.07 $\pm$ 0.09 & 113 & 131 \\
\bottomrule
\end{tabular}
\caption{External baseline comparison (void, 1,000 trials each)}
\end{table}

\subsection{Summary}

Across 63 configurations (1,000 trials each), the principal effects are large and statistically separated by non-overlapping 95\% CIs: the positive control exceeds the best ablation in spiral quality by $3.1{\times}$ (d = 1.35), the composite consumes $3.2{\times}$ more food than a single-inverter control (d = 3.51), the zero-diffusion configuration produces spiral quality 0.652 with 100\% detection, and the composite consumes 2.1-$7.9{\times}$ more food than five external baselines (all d > 4.0). All 95 prespecified validation checks are observed in the predicted direction and 92 of 95 remain significant after Holm-Bonferroni correction (the 3 non-inferred checks are exploratory H6 contrasts between adjacent configurations with d < 0.10; full test statistics in Supplementary Table S14). Twenty-one baseline configurations (21,000 additional trials) confirm that NNN advantages are specific to the circuit, not artifacts of environment geometry ({\color{refblue}\mbox{\textup{Table 14}}}).

\begin{table}[ht]
\centering
\small
\begin{tabular}{llll}
\toprule
\textbf{Hypothesis} & \textbf{Checks} & \textbf{Result} & \textbf{Key Finding} \\
\midrule
H0: Baseline & 14/14 & PASS & Single inverter = circles; speed and radius scale as predicted \\
H1: Emergence & 8/8 & PASS & 3+ inverters $\to$ spiral; quality increases with inverter count \\
H2: Specificity & 11/11 & PASS & Both opposition and staggering required; $3.1{\times}$ gap \\
H3: Exploitation & 9/9 & PASS & Spiral $\to$ tracking transition; composite eats $3.2{\times}$ more \\
H4a: Opposition & 12/12 & PASS & Near-balance = peak quality; balance = max area \\
H4b: Robustness & 5/5 & PASS & Graceful degradation; detectable at 30\% jitter \\
H5: Geometry & 7/7 & PASS & Open spiral eats 63.6\% more than tight \\
H6: Complexity & 6/6 & PASS & 3 inverters eat 41.3\% more than 7 \\
H7: Sensitivity & 5/5 & PASS & Structural proof at $D_{\mathrm{rot}}$ = 0 (quality 0.652) \\
H8: Reconciliation [exploratory] & 5/5 & PASS & 7-inv quality $2.66{\times}$ 3-inv; reverse of H6 \\
H9: Negative & 13/13 & PASS & All 11 adversarial configs fail appropriately \\
Baselines & 21 configs & -- & NNN eats 2.1-$7.9{\times}$ baselines; d > 4.0 \\
\textbf{Total} & \textbf{95/95} & \textbf{PASS} & \textbf{84,000 trials} \\
\bottomrule
\end{tabular}
\caption{Hypothesis validation summary}
\end{table}

\section{Discussion}

\subsection{What We Found}

The results establish that exploration-exploitation switching is a genuine emergent property of composite spiking oscillators rather than an artifact of parameter tuning. The spiral requires both counter-phase opposition and temporal staggering (H2), persists at zero rotational diffusion with quality 0.652 and 100\% detection rate (H7), and supports autonomous transition from search to tracking after food encounter (H3).

The H5 result shows that spiral geometry affects more than aggregate foraging efficiency. Tight spirals remain confined to a single food band ({\color{refblue}\mbox{\textup{Fig. 7a}}}), whereas open spirals sweep widely enough to discover and exploit both bands ({\color{refblue}\mbox{\textup{Fig. 7b}}}). In this sense, geometry determines how much environmental structure the agent can access, not just how much food it consumes.

\begin{figure}[H]
\centering
\begin{subfigure}[t]{0.48\textwidth}
\centering
\includegraphics[width=\textwidth]{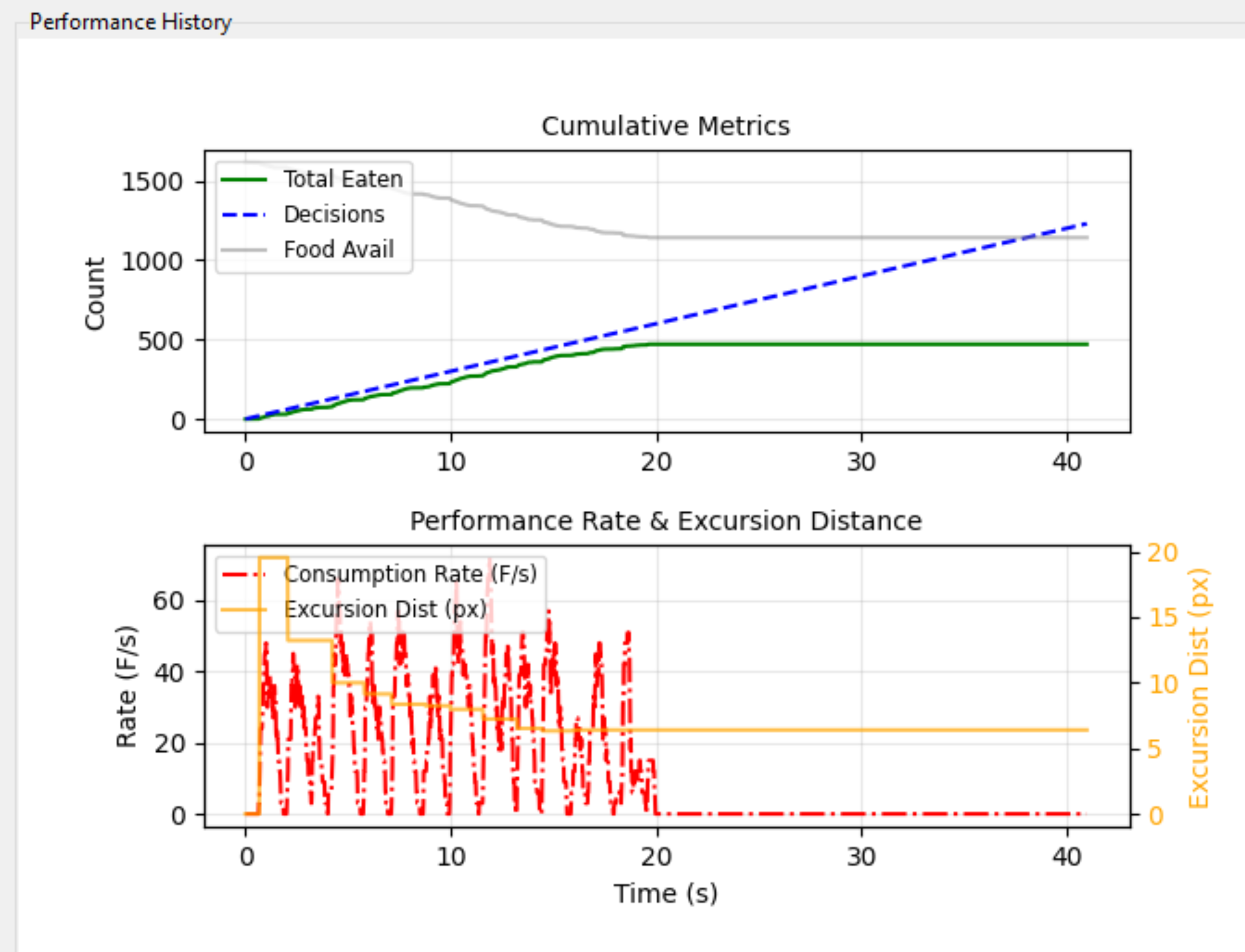}
\caption{H5 tight spiral performance: single sustained consumption period, one band only.}
\end{subfigure}
\hfill
\begin{subfigure}[t]{0.48\textwidth}
\centering
\includegraphics[width=\textwidth]{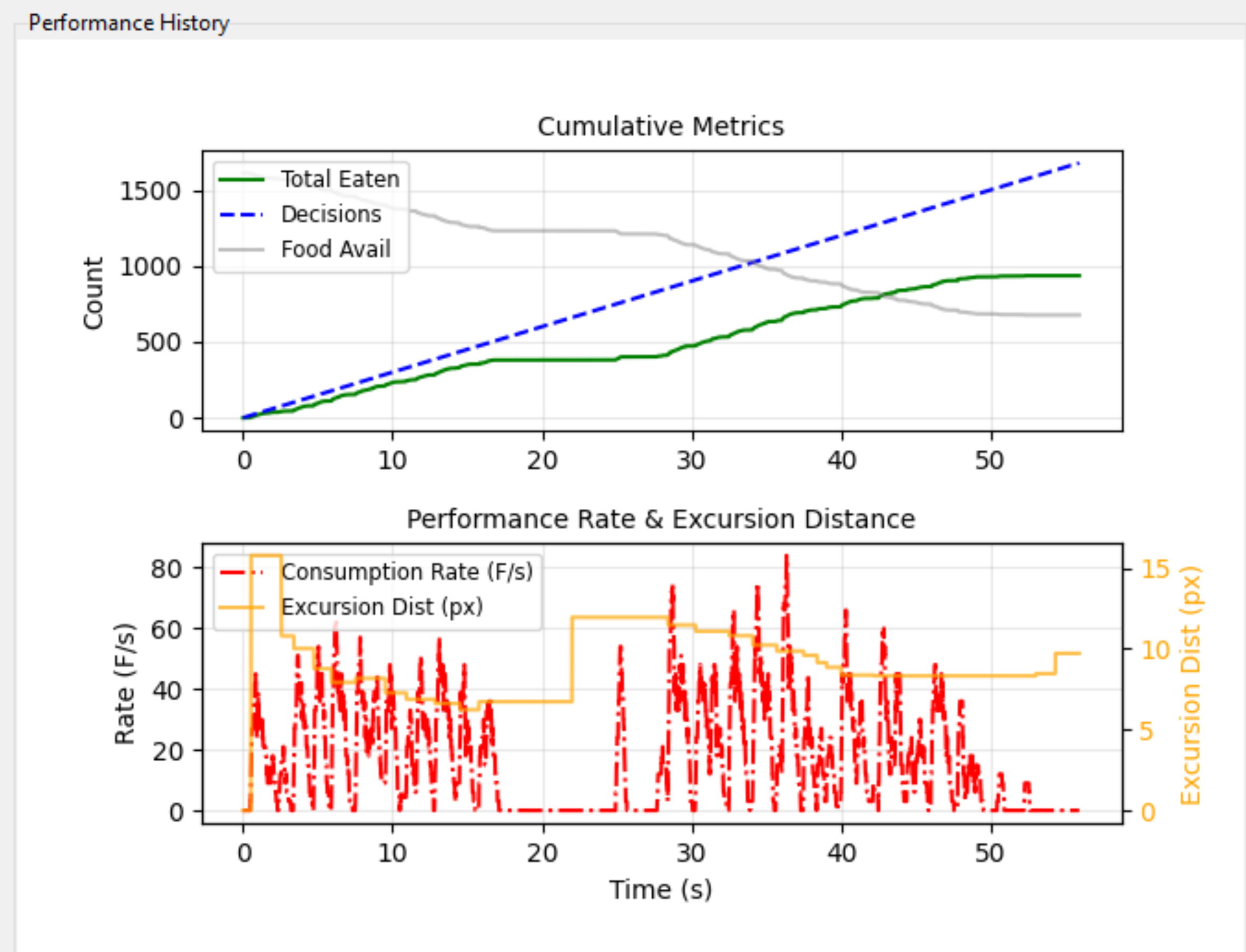}
\caption{H5 open spiral performance: double peak in consumption rate marks the discovery of the second band.}
\end{subfigure}
\caption{H5 performance traces: tight spiral (single band) vs.\ open spiral (both bands).}
\end{figure}

This role distinguishes the composite spiral from prescribed robotic spiral search. There the spiral is the solution path {\color{refblue}\cite{ref20}}. Here it is only the exploratory preamble: the useful behavior is the autonomous transition from resource discovery to continued exploitation of a structured source. Search and tracking are therefore two phases of one dynamical process rather than two separately programmed behaviors.

The mechanism is also robust. Net opposition controls trajectory character continuously (H4a), parametric perturbation degrades the spiral smoothly rather than catastrophically (H4b), and even heavy rotational noise does not eliminate the spiral signature (H7).

The quality-efficiency dissociation sharpens the main design lesson. Increasing inverter count improves spiral regularity in void (H8) but reduces food capture in structured environments (H6). The best forager is therefore not the most regular spiral, but the smallest circuit that still yields reliable search. For this task class, minimal architectures can outperform larger ones.

\subsection{Biological Analogues of Multi-Timescale Counter-Phase Coordination}

The composite circuit is not intended as a literal model of any single biological network. The relevance of the biological comparison is narrower: several well-studied systems show that multi-timescale oscillatory coordination, frequency asymmetry, and decentralized generation of structured output are biologically real organizing principles.

The clearest examples come from the crustacean stomatogastric ganglion and locomotor CPGs. In the STG, a fast pyloric rhythm modulates a much slower gastric mill rhythm through inhibitory interaction {\color{refblue}\cite{ref21,ref22,ref23}}. In lamprey, leech, and related CPG models, phase structure depends on intrinsic frequency differences and asymmetric coupling {\color{refblue}\cite{ref24,ref25,ref26,ref27}}. Mammalian half-center models further show that rhythm generation need not be symmetric: one component can dominate while others reshape the output over longer timescales {\color{refblue}\cite{ref28,ref29,ref30}}. These systems do not instantiate the present architecture, but they support the narrower claim that frequency-separated antagonistic dynamics can organize coordinated behavior.

A related point comes from bio-inspired foraging models, where switching between search modes can emerge from temporal neural dynamics without an explicit supervisory controller {\color{refblue}\cite{ref15}}. Our circuit differs from those systems in important respects: its units are independent, binary, and non-adaptive. The biological literature therefore does not validate this exact architecture. What it does support is the plausibility of the broader principle on which the model relies.

\subsection{Evolutionary Plausibility}

For the composite oscillator to be a plausible pre-neural computational primitive, it must work approximately before it can work precisely. Two lines of evidence support that requirement.

First, the mechanism tolerates substantial parametric imprecision. When all $C_1$ values are perturbed by $\pm$10\%, the spiral retains 59.8\% of its baseline quality (H4b). At $\pm$30\% perturbation, the spiral quality still exceeds the detection threshold. The degradation is smooth rather than step-like, implying that a coarse or imprecisely assembled version of the circuit would still produce functional exploratory behavior.

Second, the mechanism remains detectable under environmental noise. Deterministic performance and noisy performance address different questions (approximate construction versus noisy operation) and both are favorable. The finding that the three-inverter circuit is also the best forager further strengthens this point: evolution need not discover a complex composite before obtaining useful exploration-exploitation switching. A minimal version is already effective. Independent support comes from work demonstrating that evolutionary algorithms can produce emergent foraging behaviors in spiking neural network controllers without explicit programming of search or coordination strategies {\color{refblue}\cite{ref31,ref32}}.

\subsection{Limitations}

The agent operates in a two-dimensional plane with idealized kinematics; embodied agents in three-dimensional environments may impose additional constraints. The input signal is binary rather than graded, the agent operates in isolation, and no multi-agent interaction or communication is modeled. The trajectory metrics were designed for this model class; although negative controls and threshold sensitivity analyses show that they discriminate the intended behaviors, they are still study-specific operational measures. A direct comparison against five standard baselines (CRW, Levy walk, log spiral, ARS, stochastic; Section 5.5) confirms that the reported advantages are specific to the NNN circuit, not artifacts of environment geometry. The present results establish a minimal synthetic mechanism for adaptive switching in simulation. They do not by themselves establish that the same architecture is implemented in biological systems.

\section{Conclusions and Future Work}

\subsection{Conclusions}

If intelligence is conventionally defined by what a system has learned, this study asked whether one of its most basic signatures, adaptive search, can arise without learning at all. The results identify a minimal temporal mechanism: three or more frequency-tuned spiking oscillators, connected in counter-phase, generate autonomous switching between exploratory spiral search and exploitative tracking. The circuit requires no runtime learning, supervision, parameter adaptation, or explicit controller, situating it at the limit of unsupervised neural architectures, where all adaptive capability arises from structure rather than from a learning rule. We view it as evolutionary trained: its behavioral capabilities arise from temporal organization imposed by structural constraints that natural selection can shape without requiring experience-dependent plasticity.

The mechanism depends on multi-timescale evaluation of the same input. In the absence of signal, oscillators with different temporal windows activate sequentially, producing structured exploratory motion. When a signal appears, transient disagreement between fast and slow evaluators drives the transition from exploration to exploitation; convergence between evaluators stabilizes the new behavioral mode.

Systematic ablation establishes that the behavior is structural rather than incidental. Across 63 configurations with 1,000 independent trials each, spiral search required both temporal staggering and counter-phase opposition; neither condition alone was sufficient. The spiral persisted at zero rotational diffusion, showing that it is not a noise artifact. The mechanism also tolerated parameter perturbation and operated across a continuous range of opposition strengths, supporting its interpretation as a robust dynamical principle rather than a tuned special case.

A further result was the dissociation between trajectory regularity and adaptive efficiency. Additional oscillators improved spiral regularity in empty environments but reduced resource exploitation in resource-containing environments. Thus, more regular trajectories were not necessarily more adaptive trajectories. The most efficient agent was the minimal circuit that retained sufficient temporal structure to search, encounter, and redirect. Adaptive performance therefore depended not on circuit size alone, but on the balance between temporal differentiation and counter-phase opposition. This inverts the "bigger is better" assumption that drives current AI scaling: here, minimal circuits outperform complex ones.

We propose that the composite oscillator represents an "atomic intelligence" in a strictly operational sense: a minimal dynamical unit capable of initiating search, extending search area, and switching between exploration and exploitation autonomously. Its defining feature is proactivity: unlike conventional neural networks, which require input to produce output, the composite oscillator generates structured behavior precisely because input is absent. Intelligence, in this framing, is not reaction to stimulus but periodic action triggered by silence. This does not imply cognition, representation, or subjective experience, but at the same time provides a low-level adaptive primitive from which more complex behaviors may be composed.

The underlying logic is not unique to this model. Any system that evaluates the same signal at multiple timescales and acts on their disagreement implements a version of this principle - and this logic feels familiar because it is not limited to the model. We recognize it in our daily actions: in the way we pursue goals, expand the search rapidly in an outward spiral, then slow down and reverse direction when we notice something worth exploring. Is it not so? Intelligence was never a human prerogative; it is a feature of all living beings, and its basic principles are simpler, older, and more universal than we usually assume.

\subsection{Future Work}

Several directions follow naturally:

\begin{enumerate}
\item \textbf{Runtime adaptation.} The current study uses fixed $C_1 - C_4$ period parameters; in practice, optimal periods are environment-specific, determined by signal density, event spacing, and input reliability. A learning rule that tunes these periods in runtime (lengthening decision windows in sparse signal environments, shortening them in dense ones) would allow the circuit to adapt its exploration-exploitation balance to the statistics of its operating context, bridging the gap between the untrained foundation demonstrated here and a fully adaptive system.
\item \textbf{Evolutionary recovery of structure.} A natural next step is to ask whether the hand-specified structural conditions identified here can themselves be recovered by evolution. In such a setting, temporally structured environments would serve as implicit specifications, selection would act as the test harness, and the evolved population distribution would be the artifact to be decoded. Prior evolutionary work on spiking foraging controllers has relied on explicit fitness functions {\color{refblue}\cite{ref31,ref32}}.
\item \textbf{Associative model.} The dual-input ($2{\times}2$) extension, detailed in Appendix B, accepts two input channels and could enable temporal association, the computational analogue of classical conditioning. Associative learning has been observed in organisms as simple as cnidarians {\color{refblue}\cite{ref33}}, suggesting that this extension is biologically grounded. Earlier simulations in {\color{refblue}\cite{ref18}} introduced this $2{\times}2$ architecture and showed it to be significantly more effective than the $1{\times}2$ at resource capture.
\item \textbf{Hierarchical composition.} Connecting composites in series (where the output of one composite serves as the input to another) could implement switching at multiple spatial and temporal scales, producing nested exploration-exploitation cycles analogous to hierarchical decision-making.
\item \textbf{Hardware implementation.} Appendix A demonstrates that the translation to standard spiking network components is straightforward. Implementing the composite oscillator on neuromorphic chips would test whether the mechanism's noise tolerance extends to physical hardware.

\end{enumerate}
\subsection{Data and Code Availability}

All source code, test specifications, validation engine, figure-regeneration scripts, and package dependencies are released at \url{https://github.com/FoundAItion-ai/Its/tree/v1.1.7/PY/VisualCube} (tag v1.1.7, commit 6626c1d); the repository also includes a file manifest with SHA-256 checksums and data-to-figure mapping (MANIFEST.md). The complete dataset (raw simulation logs, per-trial metrics, and aggregated statistics for all 84,000 trials across 84 configurations) is archived at \url{https://doi.org/10.5281/zenodo.19264202.} Per-trial random seeds were not recorded; each trial used Python's default OS-seeded random state (Section 4.1), so exact stochastic re-runs from code alone are not possible. The archived per-trial outputs enable full verification of all reported statistics without re-running simulations.

\subsection{Author Contributions}

A.F. conceived the study. A.F. and M.Y. designed the simulations and analysis framework. M.Y. implemented the simulation pipeline. S.S. performed review and provided general guidance. All authors interpreted the results, wrote and revised the manuscript, approved the final version, and agree to be accountable for all aspects of the work.

\subsection{Funding}

This work received no external funding.

\subsection{Competing Interests}

The authors are affiliated with FoundAItion Inc., a non-profit research organization. The authors declare no competing interests.

\subsection{Ethics Statement}

This study involved computational simulations only and did not involve human participants, animals, or personal data. Generative AI tools were used for drafting assistance during manuscript preparation. All content was reviewed, verified, and revised by the human authors, who take full responsibility for the accuracy and integrity of the work.

\section*{Appendix A: Translation to a Neural Network Model}

The composite oscillator described in this paper is a behavioral model; it defines what the circuit does, not how it is physically implemented. This appendix shows how the model translates into a neural network architecture. The translation is relatively straightforward and follows directly from the circuit's temporal structure.

The composite inverter consists of N oscillatory elements, each tuned to a different intrinsic frequency. In the behavioral model, these are abstract inverters with period parameters $C_1 - C_4$. In a neural implementation, each element becomes a self-excitatory neuron with a feedback loop that sustains oscillation at a characteristic period $\delta t$ ({\color{refblue}\mbox{\textup{Fig. 8}}}). The key constraint is that the oscillator periods must be pairwise prime (they must not share common factors) so that the combined cycle length equals the product of all individual periods rather than their least common multiple. This requirement is critical: if periods shared a common factor, the output spikes would coincide at the least common multiple rather than the full product, shortening the effective silence detection window. By enforcing coprime periods, a small number of fast, inexpensive oscillators can generate arbitrarily long combined cycles. We term this the \textit{camerton} (tuning fork) principle {\color{refblue}\cite{ref18}}: just as a bank of tuning forks with pairwise prime frequencies produces a unique combined resonance, the composite oscillator bank produces a unique combined silence detection period. Each oscillatory neuron resonates at its own characteristic period, and the summarizing neuron S fires only when the output spikes from all N oscillators coincide temporally ({\color{refblue}\mbox{\textup{Fig. 8}}}). Because the periods are coprime, this coincidence occurs only once per combined cycle, a period equal to the product of all individual oscillator periods. Three neurons with periods of 2, 3, and 5 time units, for example, produce a combined cycle of 30 time units.

\begin{figure}[ht]
\centering
\begin{tikzpicture}[
    neuron/.style={circle, draw, very thick, minimum size=1.0cm,
                   font=\scriptsize, inner sep=0pt, align=center},
    arr/.style={->, very thick, >=stealth},
    inh/.style={->, thick, >=stealth, color=blue!70!black},
    plus/.style={font=\large\sffamily\bfseries, color=red!70!black,
                 inner sep=0pt, outer sep=0pt},
    minus/.style={font=\LARGE\sffamily\bfseries, color=blue!70!black,
                  inner sep=0pt, outer sep=0pt},
    sloop/.style={->, thick, >=stealth},
  ]
  \node[neuron] (n1) at (0, 3.0)  {Neuron\\ 1};
  \node[neuron] (n2) at (0, 0)    {Neuron\\ 2};
  \node[font=\Large] at (0,-1.5)  {$\vdots$};
  \node[neuron] (n3) at (0,-3.0)  {Neuron\\ N};

  \draw[sloop] (n1.north east) to[out=60, in=120, looseness=8] (n1.north west);
  \node[above=0.65cm of n1, font=\small] {$\Delta t_1$};
  \draw[sloop] (n2.north east) to[out=60, in=120, looseness=8] (n2.north west);
  \node[above=0.65cm of n2, font=\small] {$\Delta t_2$};
  \draw[sloop] (n3.south east) to[out=-60, in=-120, looseness=8] (n3.south west);
  \node[below=0.65cm of n3, font=\small] {$\Delta t_n$};

  \coordinate (in)    at (-4.5, 0);
  \coordinate (split) at (-3.0, 0);
  \draw[arr] (in) -- (split);

  \draw[arr] (split) -- (n1.west);
  \draw[arr] (split) -- (n2.west);
  \draw[arr] (split) -- (n3.west);
  \node[plus] at ($(n1.west)+(-0.22, 0.16)$) {+};
  \node[plus] at ($(n1.west)+(-0.22,-0.16)$) {+};
  \node[plus] at ($(n2.west)+(-0.22, 0.16)$) {+};
  \node[plus] at ($(n2.west)+(-0.22,-0.16)$) {+};
  \node[plus] at ($(n3.west)+(-0.22, 0.16)$) {+};
  \node[plus] at ($(n3.west)+(-0.22,-0.16)$) {+};

  \node[neuron] (S) at (5.0, 0) {Neuron\\ S};

  \draw[arr] (n1.east) -- (S.north west);
  \draw[arr] (n2.east) -- (S.west);
  \draw[arr] (n3.east) -- (S.south west);
  \node[plus] at ($(S.north west)+(-0.05, 0.25)$) {+};
  \node[plus] at ($(S.west)     +(-0.20, 0.22)$) {+};
  \node[plus] at ($(S.south west)+(-0.05, 0.25)$) {+};

  \draw[inh] (split) to[out=-90, in=-90, looseness=2.4]
      ($(S.south)+(-0.05, 0)$);
  \node[minus] at ($(S.south east)+(0.20,-0.15)$) {\ensuremath{-}};

  \draw[arr] (S.east) -- ++(1.2, 0);
\end{tikzpicture}
\caption{Proposed translation of the composite behavioral model into a neural network. Neurons 1 through N oscillate at different intrinsic periods ($\delta t_1$ through $\delta t_n$) via self-excitatory feedback loops. All receive the same shared input signal. Their outputs converge on a summarizing neuron S, which fires only when all input signals coincide. The inhibitory input (blue minus) from the shared signal resets the counting cycle on each input event.}
\end{figure}
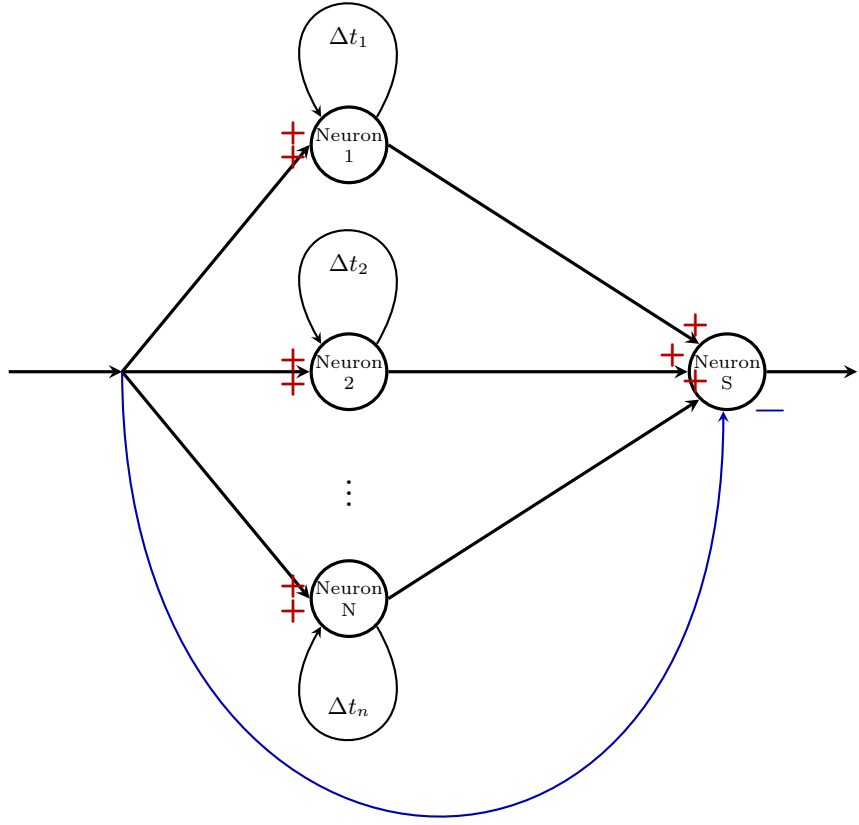

This architecture incorporates two distinct properties:

\begin{itemize}
\item A wide range of working frequencies is covered by the same topology. A small number of neurons with short individual periods can generate arbitrarily long combined silence-detection windows, scaling multiplicatively rather than additively with the number of oscillators.

\item By applying the Hebbian Rule to certain inputs, the circuit implements the behavioral model's piecewise function. When no input is detected, the self-excitatory loop takes priority: each neuron oscillates freely at its intrinsic period, and the combined output frequency is the product of all circuit frequencies. When input is detected, the shared signal takes priority: every input event restarts the counting process, producing an indefinite delay in the combined output for as long as input persists.

\end{itemize}
This architecture has a direct implication for the nature of intelligence. The composite circuit does not respond to the presence of a signal; it responds to the \textit{absence} of signals within a temporal scope. Intelligence, in this framing, is a periodic response to the absence of signals within a temporal window whose duration is determined by the combined cycle of the oscillator bank. A single neuron can only detect the absence of input over a short window (its own period). A composite of N neurons detects absence over a window equal to the product of their periods, an exponentially longer temporal horizon from a linear increase in components. This is the mechanism by which a small circuit achieves multi-scale temporal integration: each additional oscillator extends the temporal reach of the silence detector, not additively, but multiplicatively.

Thus, this network fragment can fulfill the requirements of both the $1{\times}2$ and composite N${\times}$1${\times}$2 models, serving as a basis for constructing a topologically uniform spiking neural network that describes the behavior of organisms trained by evolution to possess natural computational capabilities. Appendix B shows how this architecture extends to the $2{\times}2$ associative model, accommodating learned temporal associations within the same spiking framework.

\section*{Appendix B: The $2{\times}2$ Associative Inverter Model}

The $1{\times}2$ inverter used throughout this study accepts a single input and produces two outputs (Left and Right). In {\color{refblue}\cite{ref18}}, a $2{\times}2$ extension was introduced: the inverter accepts two inputs and produces two outputs ({\color{refblue}\mbox{\textup{Fig. 9}}}). Both inputs are of the same type but are bound to temporally offset signals, separated by a delay $\Delta$.

\begin{figure}[ht]
\centering
\begin{tikzpicture}[
    inv/.style={circle, draw, very thick, minimum size=1.2cm, font=\large\itshape},
    input/.style={->, very thick, >=stealth},
    output/.style={->, thick, >=stealth, color=black!45},
  ]
  \node[inv] (f) {$f$};

  \coordinate[left=2.5cm of f, yshift=0.35cm] (in1);
  \coordinate[left=2.5cm of f, yshift=-0.35cm] (in2);
  \draw[input] (in1) node[left, font=\small] {$I_1$} -- ([yshift=0.15cm]f.west);
  \draw[input] (in2) node[left, font=\small] {$I_2$} -- ([yshift=-0.15cm]f.west);

  \coordinate[right=2.5cm of f, yshift=0.35cm] (out1);
  \coordinate[right=2.5cm of f, yshift=-0.35cm] (out2);
  \draw[output] ([yshift=0.15cm]f.east) -- (out1) node[right, black, font=\small] {$O_1$};
  \draw[output] ([yshift=-0.15cm]f.east) -- (out2) node[right, black, font=\small] {$O_2$};
\end{tikzpicture}
\caption{The $2{\times}2$ associative inverter. Two input signals (black arrows) are evaluated within the same decision window but offset by $\Delta$. Two output signals (grey arrows) are produced. The function f determines the output periods based on the joint input state.}
\end{figure}
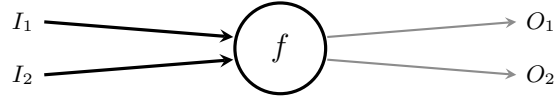

The output periods are determined by the joint state of both inputs within the decision window. As in Section 3, the windowed event-count formulation is used. Let $N_1$ and $N_2$ denote the event counts from the two input channels during the decision window:

\begin{equation}
f_{o1} = \begin{cases} C_1, & N_1(t,\,t{+}C_1) = 0 \text{ and } N_2(t{+}\Delta,\,t{+}\Delta{+}C_1) = 0, \\
C_2, & N_1(t,\,t{+}C_1) \geq 1 \text{ and } N_2(t{+}\Delta,\,t{+}\Delta{+}C_1) \geq 1, \end{cases}
\end{equation}
\begin{equation}
f_{o2} = \begin{cases} C_3, & N_1(t,\,t{+}C_1) = 0 \text{ and } N_2(t{+}\Delta,\,t{+}\Delta{+}C_1) = 0, \\
C_4, & N_1(t,\,t{+}C_1) \geq 1 \text{ and } N_2(t{+}\Delta,\,t{+}\Delta{+}C_1) \geq 1. \end{cases}
\end{equation}
where $N_1$ and $N_2$ are the event counts from the two input channels within their respective decision windows, and $C_1$ through $C_4$ are the same period parameters as in the $1{\times}2$ model.

The critical feature is the temporal offset $\Delta$ between the two inputs. The inverter fires at its active-state periods ($C_1$, $C_3$) only when \textit{both} inputs are absent within the decision window; that is, when both signals are absent within their respective temporal scopes. If the first input arrives and the second follows within $\Delta$, the inverter learns to associate the two signals: the first becomes a predictor of the second. This is structurally equivalent to classical conditioning, in which a conditioned stimulus (first input) acquires predictive value for an unconditioned stimulus (second input) through repeated temporal pairing.

Associative learning has been observed in organisms as simple as cnidarians: \textit{Nematostella vectensis}, a sea anemone with no centralized nervous system, demonstrates learned associations between light and electric shock {\color{refblue}\cite{ref33}}. This suggests that the temporal-pairing mechanism implemented by the $2{\times}2$ inverter may have biological precedent at the earliest stages of neural evolution. Empirically, {\color{refblue}\cite{ref18}} reported that the $2{\times}2$ architecture produced a foraging behaviour pattern very similar to the $1{\times}2$ but with significantly higher effectiveness. This gain provides quantitative motivation independent of the biological argument. The $2{\times}2$ model thus represents a natural next step: augmenting the composite oscillator's exploration-exploitation switching with the capacity for learned temporal associations, without departing from the spiking oscillator architecture.

\end{document}